\documentclass[10pt,twocolumn]{article}

\usepackage{graphicx}
\usepackage{amsmath}
\usepackage{amssymb}
\usepackage[latin1]{inputenc}
\usepackage{cite}
\usepackage{mathtools}
\usepackage{multirow}
\usepackage{xcolor}

\begin{document}

\title{Salient Object Detection by LTP Texture\\ 
       Characterization on Opposing Color Pairs\\ 
        under SLICO Superpixel Constraint}

\author{Didier Ndayikengurukiye and Max Mignotte
    \thanks{The authors are with the vision lab. of the D\'epartement
    d'Informatique et de Recherche Op\'erationnelle (DIRO),
    Universit\'e de Montr\'eal, Facult\'e des Arts et des Sciences,
    Montr\'eal, H3C 3J7, QC, Canada. E-mails: 
     didier.ndayikengurukiye@umontreal.ca, \, mignotte@iro.umontreal.ca}}

\maketitle

\begin{abstract}
The effortless detection of salient objects by humans has been the
subject of research in several fields, including computer vision as it
has many applications.  However, salient object detection remains a
challenge for many computer models dealing with color and textured
images.  Herein, we propose a novel and efficient strategy, through a
simple model, almost without internal parameters, which generates a
robust saliency map for a natural image.  This strategy consists of
integrating color information into local textural patterns to
characterize a color micro-texture. Most models in the literature that
use the color and texture features treat them separately. In our case,
it is the simple, yet powerful LTP (Local Ternary Patterns) texture
descriptor applied to opposing color pairs of a color space that
allows us to achieve this end.  Each color micro-texture is
represented by vector whose components are from a superpixel obtained
by SLICO (Simple Linear Iterative Clustering with zero parameter)
algorithm which is simple, fast and exhibits state-of-the-art boundary
adherence.  The degree of dissimilarity between each pair of color
micro-texture is computed by the FastMap method, a fast version of MDS
(Multi-dimensional Scaling), that considers the color micro-textures
non-linearity while preserving their distances. These degrees of
dissimilarity give us an intermediate saliency map for each RGB, HSL,
LUV and CMY color spaces. The final saliency map is their combination
to take advantage of the strength of each of them.  The MAE (Mean
Absolute Error) and F$_{\beta}$ measures of our saliency maps, on
the complex ECSSD dataset show that our model is both simple and
efficient, outperforming several state-of-the-art models.\\
\\


\end{abstract}

\vspace{0ex}
\section{Introduction}
\label{Introduction}

Humans - or animals in general -  have a visual
system endowed with attentional mechanisms. These mechanisms allow the
human visual system (HVS) to select from the large amount of
information received that which is relevant and to process in detail
only the relevant one \cite{parkhurst2002modeling}. This phenomenon is
called \textit{visual attention}. This mobilization of resources for
the processing of only a part of whole information allows its rapid
processing. Thus the gaze is quickly directed towards certain objects
of interest. For living beings, this can sometimes be vital as they
can decide whether they are facing a prey or a predator
\cite{itti2005models}.

Visual attention is carried out in two ways, namely \textit{bottom-up
  attention} and \textit{top-down attention
}\cite{itti2001computational}. \textit{Bottom-up attention } is a
process which is fast, automatic, involuntary, directed by the image
properties almost exclusively \cite{parkhurst2002modeling}. The
\textit{top-down attention} is a slower, voluntary mechanism directed
by cognitive phenomena as knowledge, expectations, rewards, and
current goals \cite{baluch2011mechanisms}. In this work, we focus on
the \textit{bottom-up attentional mechanism} which is image-based.\\
Visual attention has been the subject of several research works in the
fields of cognitive psychology \cite{treisman1988features,
  wolfe1989guided}, neuroscience \cite{koch1987shifts}, to name a
few. Computer vision researchers have also used the advances in
cognitive psychology and neuroscience to set up computational visual
saliency models that exploit this ability of the human visual system
to quickly and efficiently understand an image or a scene. Thus, many
computational visual saliency models have been proposed and for more
details, most of the models can be found in these works
\cite{borji2019salient, borji2012state}. Computational visual saliency
models are mainly oriented eye fixation prediction and salient objects
segmentation or detection. The latter is the subject of this work.\\

Computational visual saliency models have several
applications. Indeed, saliency models in multimedia are used in
\textit{image/video compression} \cite{itti2004automatic}, \textit{image correction} \cite{li2020saliency},
\textit{image retrieval}\cite{gao2015database}, \textit{advertisements
  optimization} \cite{pieters2004attention}, \textit{aesthetics
  assessment} \cite{wong2009saliency}, \textit{image quality assessment}
\cite{liu2009studying}. Saliency models are also used in \textit{image
  retargeting} \cite{chen2003visual}, \textit{image montage}
\cite{chen2009sketch2photo}, \textit{image collage}
\cite{yang2022visual, huang2011arcimboldo}, \textit{object recognition, tracking, and
  detection}  \cite{smeulders2013visual}, \ldots.
  
The salient object detection is materialized by saliency maps or by
tracing boxes around the salient objects. In this work, we
estimate saliency maps.  The \textit{saliency map}, for an observed
image, highlights the salient objects while considering the other
objects which are not salient as background. Concretely, a saliency
map is represented by a grayscale image in which a pixel must be
whiter as it probably corresponds to a salient zone in the sense of
the human visual system. It means that this pixel is more dissimilar
than the other pixels of the image in terms of texture, color, shape,
gradient distribution or generally any attribute perceived by the
human visual system. Thus, the area of interest chosen by the human
visual system generally corresponds to a shape, a set of shapes with a
color, a mixture of colors, a movement or a discriminating texture in
the scene which differs significantly from the rest of the image.

Herein, we propose a simple and nearly parameter-free model which
gives us an efficient saliency map for a natural image using a new
strategy. The proposed model uses texture and color features in a way
that integrate color in texture feature using simple and efficient
algorithms. Contrary to classical salient detection methods, we took
the {\em texture} as an essential feature to give us the information
we need to obtain a saliency map of an image. Indeed, the {\em texture} is a ubiquitous phenomenon in natural images : images of
mountains, trees, bushes, grass, sky, lakes, roads, buildings,
etc. appear as different types of texture\footnotemark[1]
\cite{pietikainen2011computer}. In addition, natural images are
usually also color images and it is then important to take this factor
into account as well. In our application, the color is taken into
account and integrated in an original way, {\em via} the extraction
of the textural characteristics made on the pairs of opposing color
spaces.
%
\footnotetext[1]{
\samepage 
Haidekker \cite{haidekker2011advanced} argues that {\em texture} and
shape analysis are very powerful tools for extracting image
information in an unsupervised manner. This author adds that the {\em
  texture} analysis has become a key step in the quantitative and
unsupervised analysis of biomedical images
\cite{haidekker2011advanced}. Other authors, such as Knutsson and
Granlund \cite {knutsson1983texture}, Ojala {\em et al.}
\cite{ojala1996comparative},  agree that {\em texture} is an important
feature for scene analysis of images. Knutsson and Granlund also claim
that the presence of a {\em texture} somewhere in an image is more a
rule than an exception. Thus, {\em texture} in the image has been
shown to be of great importance for image segmentation, interpretation
of scenes \cite{laws1980textured}, in face recognition, facial
expression recognition, face authentication, gender recognition, gait
recognition, age estimation, to just name a few
\cite{pietikainen2011computer}.
}

Although there is a lot of work relating to {\em texture}, there is no
formal definition of {\em texture} \cite{knutsson1983texture}. There is also no agreement on a single
technique for measuring texture \cite{laws1980textured,
  pietikainen2011computer}. Our model uses the LTP ({\em local ternary
  patterns}) \cite{tan2010enhanced} texture measurement technique. The
LTP (local ternary patterns) is an extension of local binary pattern
(LBP) with 3 code values instead of 2 for LBP. LBP is known to be a powerful texture descriptor \cite{pietikainen2011computer}. Its main qualities are invariance against
monotonic gray level changes and computational simplicity and its
drawback is that it is sensitive to noise in uniform regions of the
image. In contrast, LTP is more discriminant and less sensitive to
noise in uniform regions. The LTP ({\em Local Ternary Patterns}) is
therefore better suited to tackle our salience detection
problem. Certainly, the presence in natural images of several patterns
make the detection of salient objects complex. However, the model we
propose does not just focus on the patterns in the image by processing
them separately from the colors in this image as most models do
\cite{margolin2013makes, zhang2017salient}. In this work, we propose an approach of the
salient objects detection  by taking into account both the presence in
natural images of several patterns and color not separately. This task
of integrating color in texture feature is accomplished through LTP
(Local Ternary Patterns) applied to opposing color pairs of a given
color space. The LTP describes the local textural patterns for a
grayscale image through a code assigned to each pixel of the image by
comparing it with its neighbours. When LTP is applied to an opposing
color pair, the principle is similar to that used for a grayscale
image. However, for LTP on an opposing color pair, the local textural
patterns are obtained thanks to a code assigned to each pixel, but the
value of the pixel of the first color of the pair is compared to the
equivalents of its neighbours in the second color of the pair. The
color is thus integrated to the local textural patterns. Thus, we
characterize the color micro-textures of the image without separating
the textures in the image and the colors in this same image. The color
micro-textures boundaries correspond to the superpixel obtained
thanks to the SLICO (Simple Linear Iterative Clustering with zero
parameter) algorithm which is faster and exhibits state-of-the-art
boundary adherence. A feature vector representing the color
micro-texture is obtained by the concatenation of the histograms of
the superpixel (defining the micro-texture) of each opposing color
pair. Each pixel is then characterized by a vector representing the
color micro-texture to which it belongs. 

We then compare the color micro textures characterizing each pair of
pixels of the image being processed thanks to the fast version of MDS
(multi-dimensional scaling) method \textit{FastMap}. This comparison
permits to capture the degree of pixel's uniqueness or pixel's
rarity. The FastMap method will allow this capture while taking into
account the non-linearities in the representation of each
pixel. Finally, since there is no single color space suitable for
color texture analysis\cite{porebski2008haralick}, we combine the
different maps generated by FastMap from different color spaces, such
as RGB, HSL, LUV and CMY, to exploit each other's strengths in the
final saliency map. The details of this model are described in the
section \ref{sec:ProposedModel}.


\section{Related work} \label{sec:Relatedwork}

Most authors define salient objects detection as a capture of the
uniqueness, distinctiveness, or rarity of a pixel, a superpixel, a
patch, or a region of an image \cite{borji2019salient}. The problem of
detecting salient objects is therefore to find the best
characterization of the pixel, the patch or the superpixel and to find
the best way to compare the different pixels (patch or superpixel)
representation to obtain the best saliency maps. In this section, we
present some models related to this work approach with an emphasis on
the features used and how their dissimilarities are computed. 

Thanks to studies in cognitive psychology and neuroscience, such as
those by Treisman and Gelade \cite{treisman1980feature}, Wolfe {\em et
  al.}\cite{wolfe1989guided, wolfe2004attributes} and Koch and Ullman
\cite{koch1987shifts}, the authors in the seminal work of Itti {\em et
  al.} \cite {itti1998model} - oriented  eye fixation prediction -
chose as features: color, intensity and orientation. Frintrop {\em et
  al.} \cite {frintrop2015traditional} adapting the Itti {\em et al.}
model\cite {itti1998model} for salient objects segmentation - or
detection - chose  color and intensity as features. In the two latter
models, the authors used pyramids of Gaussian and center-surround
differences to capture the distinctiveness of pixels.

The Achanta {\em et al.} model \cite{achanta2008salient} and the
histogram-based contrast (HC) model \cite{cheng2015global} used color
in CIELab space to characterize a pixel. In the latter model, the
pixel's saliency is obtained using its color contrast to all other
pixels in the image by measuring distance between the pixel for which
they are computing saliency and all other pixels in the image; this is
coupled with a smoothing procedure to reduce quantization
artifacts. The Achanta {\em et al.} model \cite{achanta2008salient}
computed pixel's saliency on three scales. For each scale, this
saliency is computed as the Euclidean distance between the average
color vectors of the inner region $R_1$ and that of the outer region
$R_2$ both centered on that pixel above mentioned. 
Guo {\em and} Zhang \cite {guo2009novel} in the phase spectrum of
Quaternion Fourier Transform model represent each image's pixel by a
Quaternion  that consists of color, intensity and motion feature. A
Quaternion Fourier Transform (QFT) is then applied to that
representation of each pixel. After setting the module of the result
of the QFT to 1 to keep only the phase spectrum in the frequency domain,
this result is used to reconstruct the Quaternion in spatial
space. The module of this reconstructed Quaternion is smoothed with a
Gaussian filter and this then gives the spatio-temporal saliency map
of their model. For static images the motion feature is set to zero.\\ 
Other models also take color and position as features to characterize
a region or patch instead of a pixel\cite{cheng2015global,
  perazzi2012saliency, goferman2012context}. They differ, however, in
how they get the salience of a region or patch. Thus, the region-based
contrast (RC) model \cite{cheng2015global} measured the region
saliency as the contrast between this region and the other regions of
the image. This contrast is also weighted depending on the spatial
distance of this region relative to the other regions of the image.\\
In the Perazzi {\em et al.} model \cite {perazzi2012saliency},
contrast is measured by the uniqueness rate and the spatial
distribution of small perceptually homogeneous regions. The uniqueness
of a region is calculated as the sum of the Euclidean distances
between its color and the color of each region weighted by a Gaussian
function of their relative position. The spatial distribution of a
region is given by the sum of the Euclidean distances between its
position and the position of each region weighted by a Gaussian
function of their relative color. The region saliency is a combination
of its uniqueness and its spatial distribution. Finally, the saliency
of each pixel in the image is a linear combination of the saliency of
homogeneous regions. The weight for each region saliency of this sum
is a Gaussian function of the Euclidean distances between the color of
the pixel and the colors of the homogeneous regions and the Euclidean
distances between its spatial position and theirs. In the 
Goferman {\em et al.} model \cite {goferman2012context}, the dissimilarity
between two patches is defined as directly proportional to the
Euclidean distance between the colors of the two patches and inversely
proportional to their relative position normalized to be between 0 and
1. The salience of a pixel at a given scale is then 1 minus the
inverse of the exponential of the mean of the dissimilarity between
the patch centered on this pixel and the patches which are more
similar to it. The final saliency of the pixel being the average of
the saliency of the different scales to which they add the context.\\
Some models focus on the patterns as feature but they compute patterns
separately from colors \cite{margolin2013makes, zhang2017salient}. For example Margolin  {\em et al.} \cite{margolin2013makes} defined a salient object as consisting of pixels whose local neighbourhood (region or patch) is distinctive in both color and pattern. The final saliency of their model is the product of the color and pattern distinctness weighted by a Gaussian to add a center-prior.\\
As Frintrop {\em et al.} \cite {frintrop2015traditional} stated most saliency systems use intensity and color features. They are
differentiated by the feature extraction and the general structure of
the models. They have in common the computation of the contrast
relative to the features chosen since the salient objects are so
because the importance of their dissimilarities with their
environment. However, models in the literature differ on how these
dissimilarities are obtained. Even though there are many salient objects detection models, the detection of salient objects remains a challenge \cite{qi2015saliencyrank}. 

\vspace{1ex}
The main contribution of this work is that we propose an unexplored
approach to the detection of salient objects. Indeed, we use for the
first time in the salient object detection, to our knowledge, the
feature \textit{color micro-texture} in which the \textit{color}
feature is integrated \textit{algorithmically} into the local textural
patterns for salient object detection. This is done by applying LTP
(Local Ternary Patterns) to each of the opposing color pairs of a
chosen color space. Thus, in salient object detection computation, we
\textit{integrate} the color information in the texture while most of
the models in the literature which use these two visual features,
namely color and texture, perform this computation
\textit{separately}. 

We also  use the \textit{FastMap} method which  conceptually is both
local and global and can be seen as a nonlinear one-dimensional
reduction of the micro-texture vector taken locally around each pixel
with the interesting constraint that the (Euclidean) difference
existing between each pair of (color) micro textural vectors
(therefore centered on two pixels of the original image) is preserved
in the reduced (one-dimensional) image and represented (after
reduction) by two gray levels separated by this same distance.  After
normalization, a saliency measure map (with range values between $0$
and $1$) is estimated in which lighter regions are more salient
(higher relevance weight) and darker regions are less salient.
Most of the models in the literature use either a local approach or a
global approach and other models combine these approaches in saliency
detection. The model that we propose in this work is simple and
parameter-free yet it performs well. The model we propose in this is
both simple and efficient while being almost parameter free. In
addition, it gives good results in comparison with state-of-the-art
models in \cite{borji2015salient} for the ECSSD dataset and for
MSRA10K.

\vspace{0ex}
\section{Proposed Model}
\label{sec:ProposedModel}

\subsection{Introduction}
\label{IntroProposedModel}

The main idea of our model is to algorithmically integrate the color
feature into the textural characteristics of the image and then to
describe this vector of textural characteristics by an intensity
histogram.

To incorporate the color into the texture description, we  mainly
relied on the opponent color theory. This theory states that the HVS
interprets information about color by processing signals from the cone
and rod cells in an antagonistic manner. This theory was suggested as
a result of the way in which photo-receptors are interconnected
neurally and also by the fact that it is made more efficient for the
HVS to record differences between the responses of cones, rather than
each type of cone's individual response. The opponent color theory
suggests that there are three opposing channels called the cone
photo-receptors which are linked together to form three pairs of
opposite colors. This theory was first computer modeled for
incorporating the color into the LBP texture descriptor by
M{\"a}enp{\"a}{\"a} {\em and} Pietik{\"a}inen
\cite{maenpaa2004classification, pietikainen2011computer}. It was
called Opponent-Color LBP (OC-LBP), and was developed as a joint
color-texture operator, thus generalizing the classical LBP, which
normally applies to monochrome textures.

Our model is locally based (for each pixel) on nine opposing color
pairs and semi-locally, on the set of estimated superpixels of the
input image. These nine opposing color pairs are in the RGB (Red -
Green - Blue) color space channel : RR, RG, RB, GR, GG, GB, BR, BG and
BB (see Figure \ref{fig:opponentRGB}). 

\begin{figure}[ht]
\centering
\begin{tabular}{ccccc}
(a) &
\hspace{-2ex} \includegraphics[width=0.09\textwidth,height=0.09\textwidth]{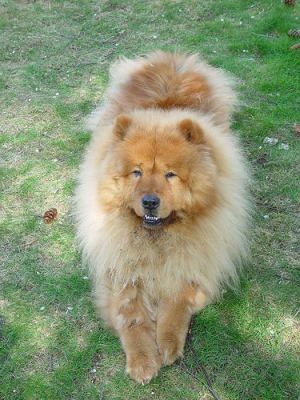}    &
\hspace{-2ex} \includegraphics[width=0.09\textwidth,height=0.09\textwidth]{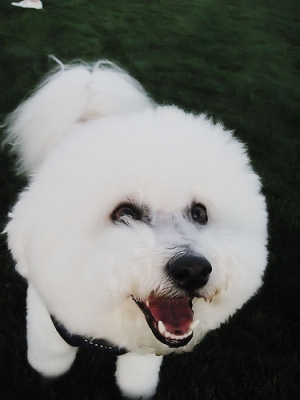}    &
 \hspace{-2ex} \includegraphics[width=0.09\textwidth,height=0.09\textwidth]{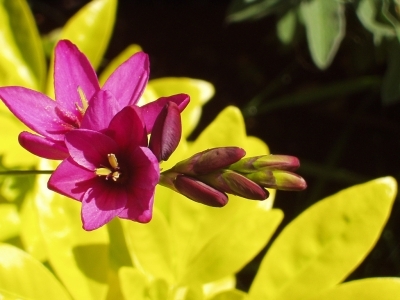} &
 \hspace{-2ex} \includegraphics[width=0.09\textwidth,height=0.09\textwidth]{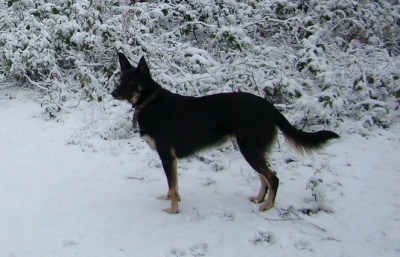} \\
 (b) &
 \hspace{-2ex}
\includegraphics[width=0.09\textwidth,height=0.09\textwidth]{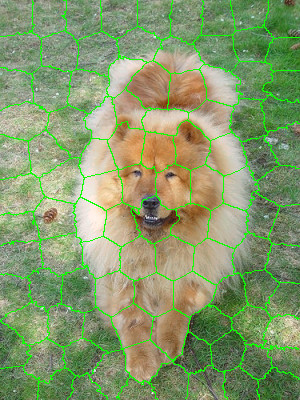}    &
 \hspace{-2ex}
\includegraphics[width=0.09\textwidth,height=0.09\textwidth]{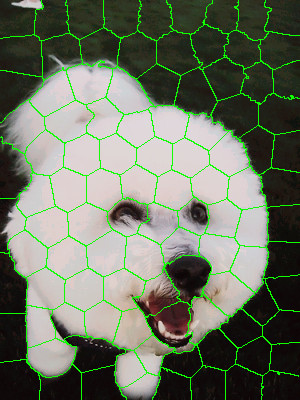}    &
 \hspace{-2ex} \includegraphics[width=0.09\textwidth,height=0.09\textwidth]{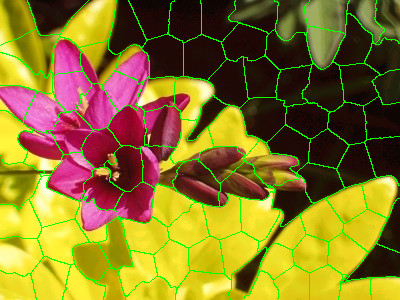} &
 \hspace{-2ex} \includegraphics[width=0.09\textwidth,height=0.09\textwidth]{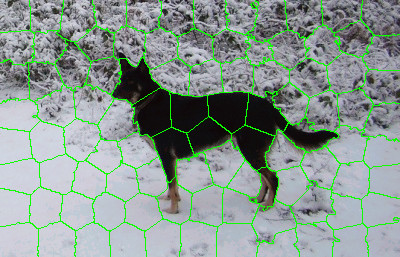} \\
\end{tabular}
\caption{\label{fig:slico_superpixels} Illustration of SLICO superpixels bounderies: (a) images ; (b) superpixels.} 
\end{figure}

\begin{figure*}[ht]
\centering
\begin{tabular}{cccccccccc}
Image & RR & RG & RB & GR & GG & GB & BR & BG & BB \\
 \hspace{-2ex} \includegraphics[scale=0.1]{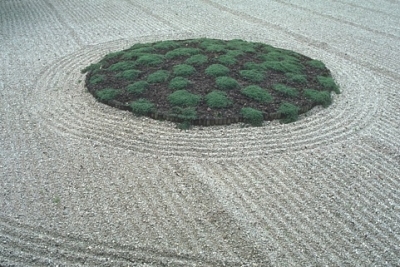}    &
 \hspace{-2ex} \includegraphics[scale=0.1]{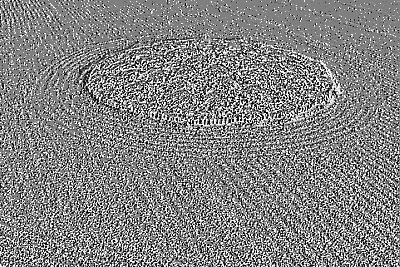} &
 \hspace{-2ex} \includegraphics[scale=0.1]{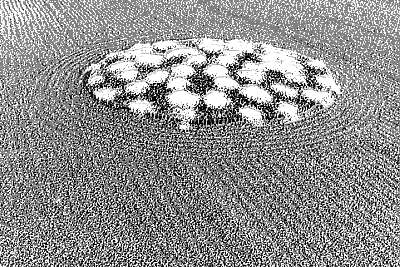} &
 \hspace{-2ex} \includegraphics[scale=0.1]{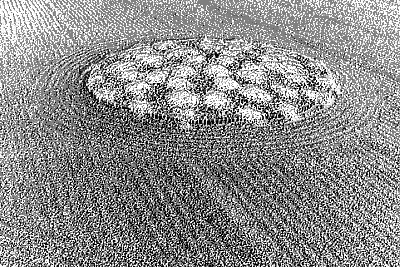} &
 \hspace{-2ex} \includegraphics[scale=0.1]{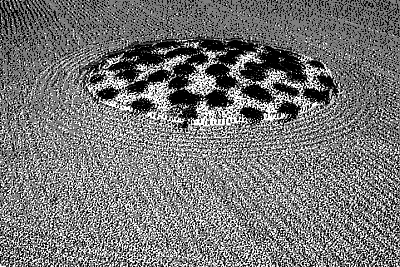} &
 \hspace{-2ex} \includegraphics[scale=0.1]{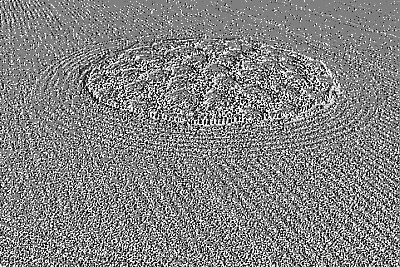} &
 \hspace{-2ex} \includegraphics[scale=0.1]{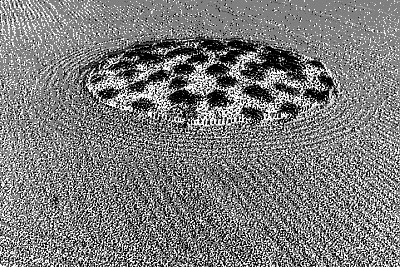} &
 \hspace{-2ex} \includegraphics[scale=0.1]{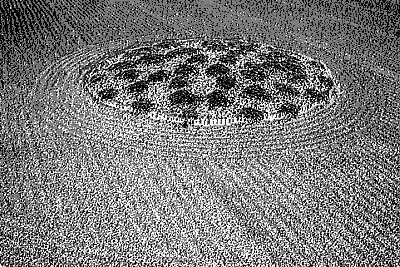} &
 \hspace{-2ex} \includegraphics[scale=0.1]{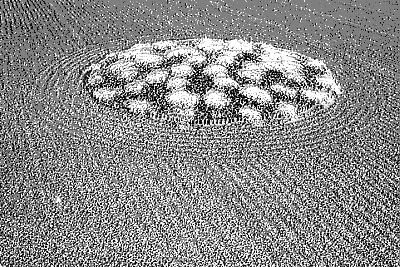} &
 \hspace{-2ex} \includegraphics[scale=0.1]{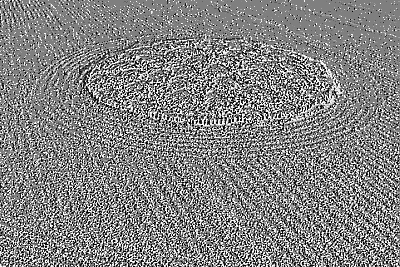} \\
 \hspace{-2ex} \includegraphics[scale=0.1]{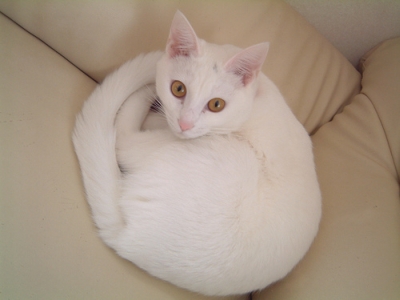}    &
 \hspace{-2ex} \includegraphics[scale=0.1]{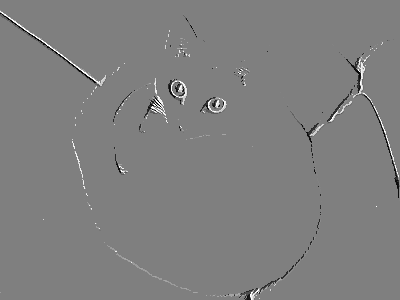} &
 \hspace{-2ex} \includegraphics[scale=0.1]{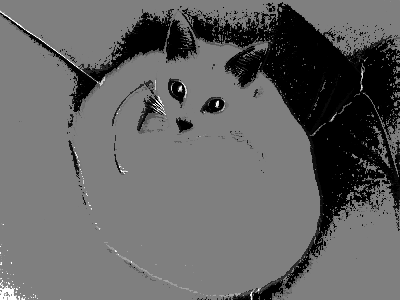} &
 \hspace{-2ex} \includegraphics[scale=0.1]{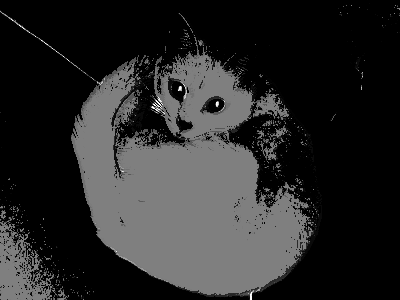} &
 \hspace{-2ex} \includegraphics[scale=0.1]{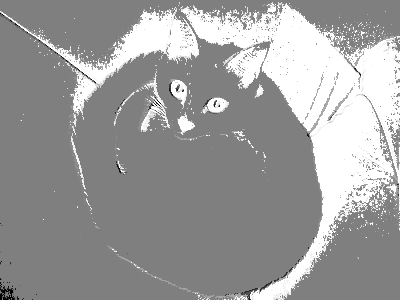} &
 \hspace{-2ex} \includegraphics[scale=0.1]{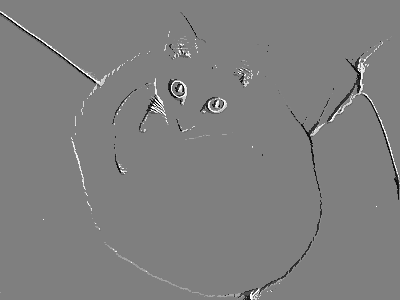} &
 \hspace{-2ex} \includegraphics[scale=0.1]{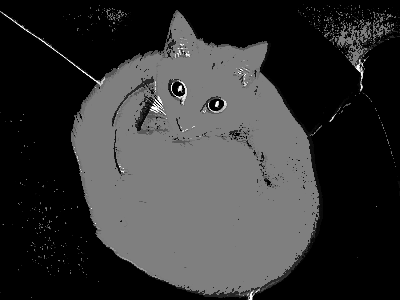} &
 \hspace{-2ex} \includegraphics[scale=0.1]{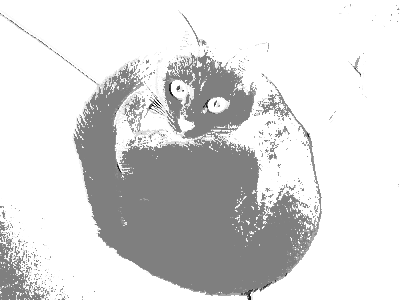} &
 \hspace{-2ex} \includegraphics[scale=0.1]{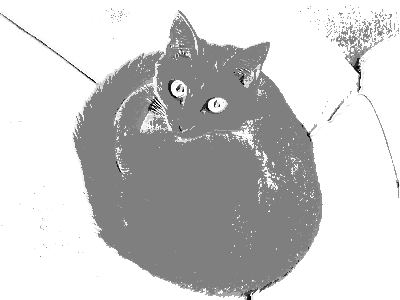} &
 \hspace{-2ex} \includegraphics[scale=0.1]{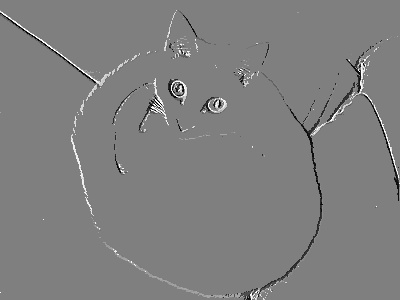} \\
 \hspace{-2ex} \includegraphics[scale=0.1]{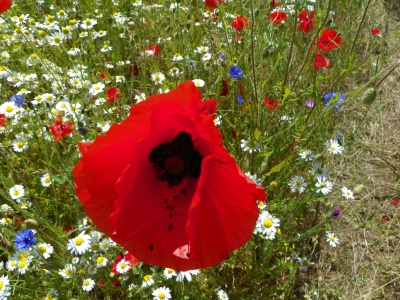}    &
 \hspace{-2ex} \includegraphics[scale=0.1]{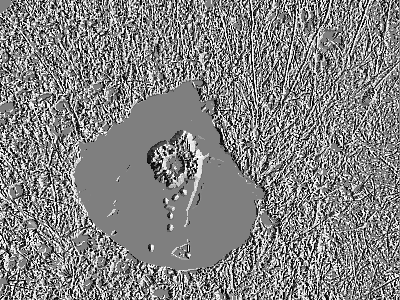} &
 \hspace{-2ex} \includegraphics[scale=0.1]{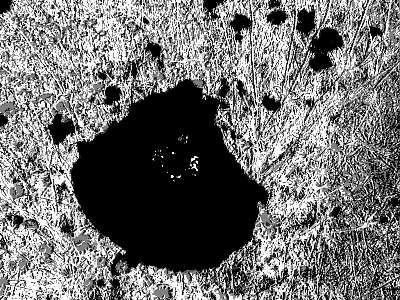} &
 \hspace{-2ex} \includegraphics[scale=0.1]{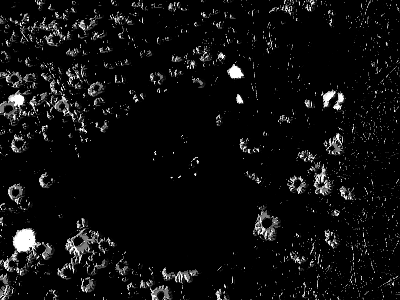} &
 \hspace{-2ex} \includegraphics[scale=0.1]{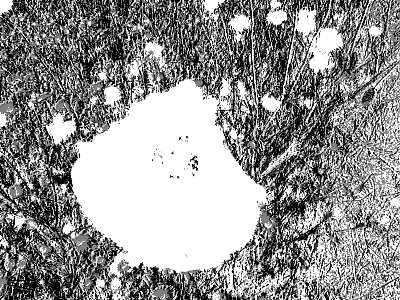} &
 \hspace{-2ex} \includegraphics[scale=0.1]{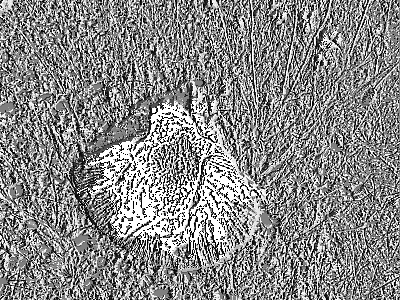} &
 \hspace{-2ex} \includegraphics[scale=0.1]{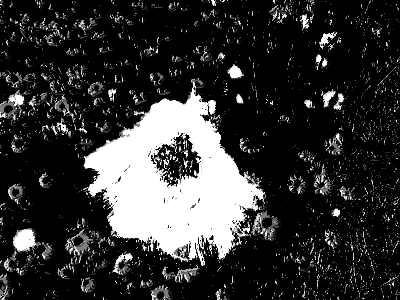} &
 \hspace{-2ex} \includegraphics[scale=0.1]{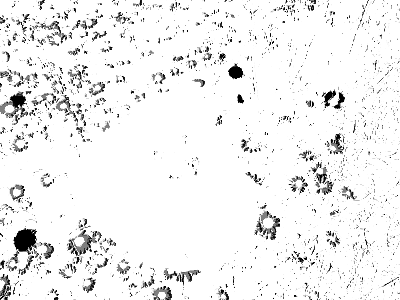} &
 \hspace{-2ex} \includegraphics[scale=0.1]{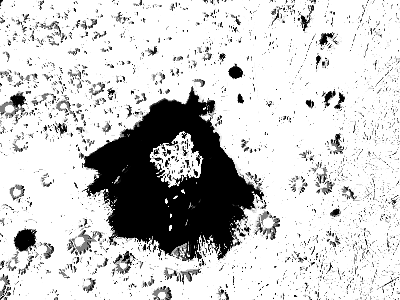} &
 \hspace{-2ex} \includegraphics[scale=0.1]{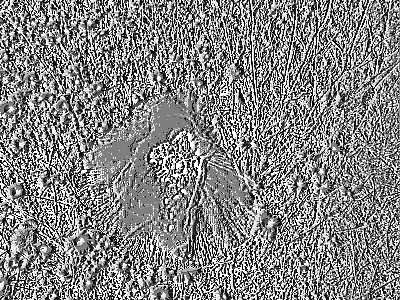} \\
 \hspace{-2ex} \includegraphics[scale=0.4]{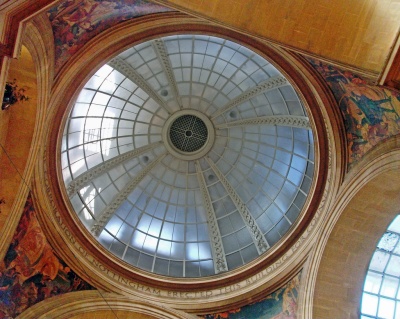}    &
 \hspace{-2ex} \includegraphics[scale=0.1]{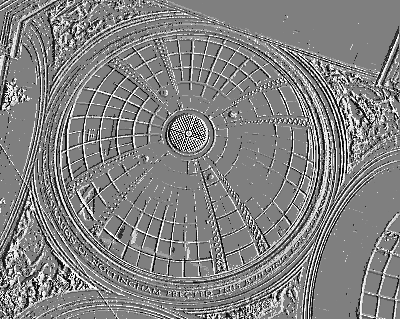} &
 \hspace{-2ex} \includegraphics[scale=0.1]{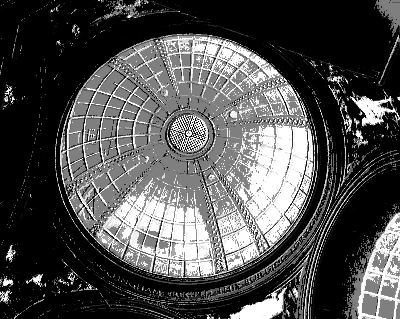} &
 \hspace{-2ex} \includegraphics[scale=0.1]{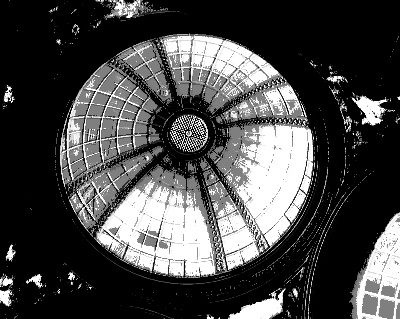} &
 \hspace{-2ex} \includegraphics[scale=0.1]{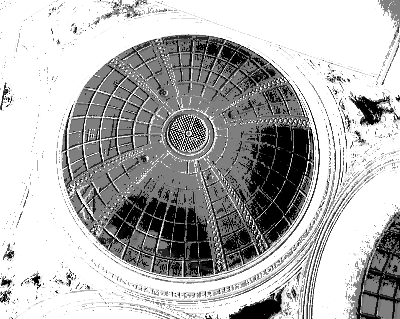} &
 \hspace{-2ex} \includegraphics[scale=0.1]{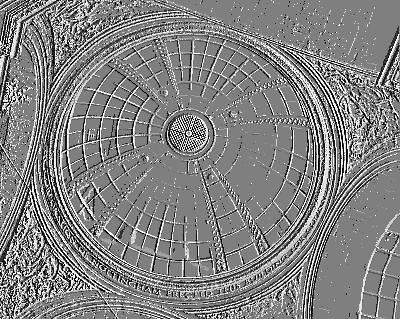} &
 \hspace{-2ex} \includegraphics[scale=0.1]{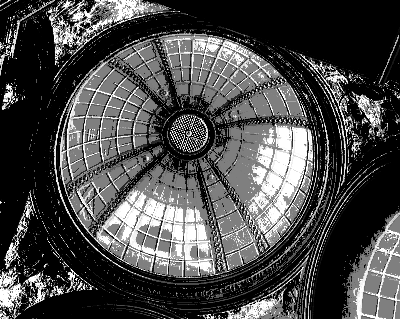} &
 \hspace{-2ex} \includegraphics[scale=0.1]{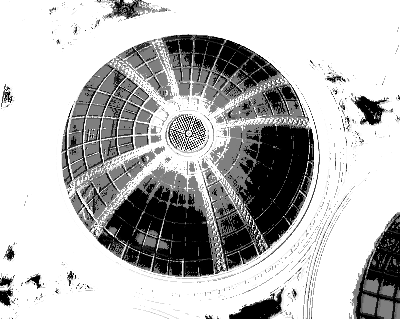} &
 \hspace{-2ex} \includegraphics[scale=0.1]{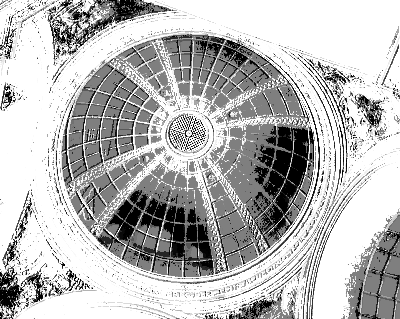} &
 \hspace{-2ex} \includegraphics[scale=0.1]{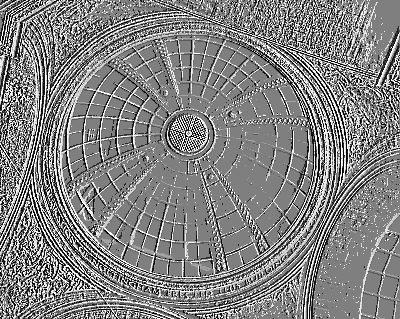} \\
 \hspace{-2ex} \includegraphics[scale=0.25]{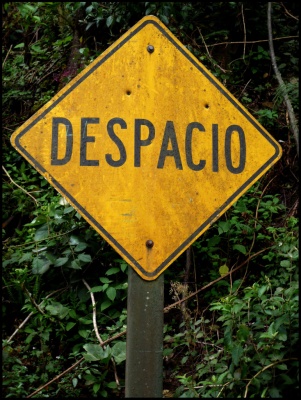}    &
 \hspace{-2ex} \includegraphics[scale=0.1]{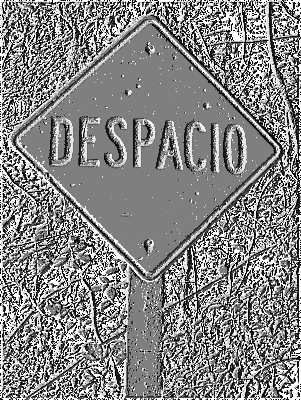} &
 \hspace{-2ex} \includegraphics[scale=0.1]{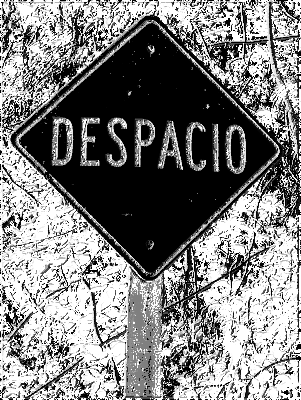} &
 \hspace{-2ex} \includegraphics[scale=0.1]{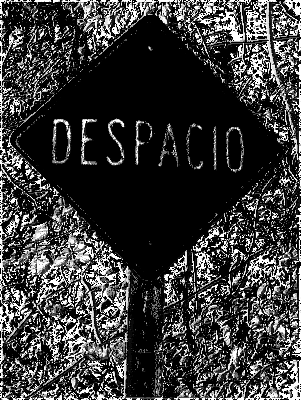} &
 \hspace{-2ex} \includegraphics[scale=0.1]{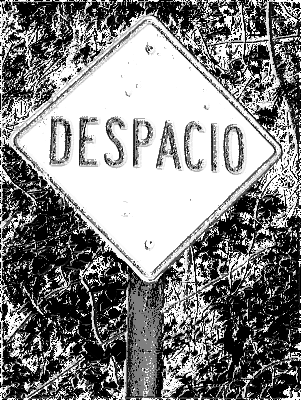} &
 \hspace{-2ex} \includegraphics[scale=0.1]{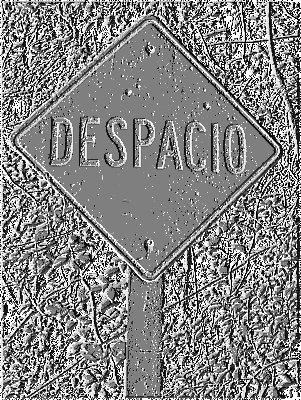} &
 \hspace{-2ex} \includegraphics[scale=0.1]{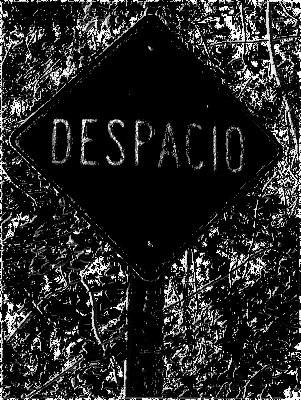} &
 \hspace{-2ex} \includegraphics[scale=0.1]{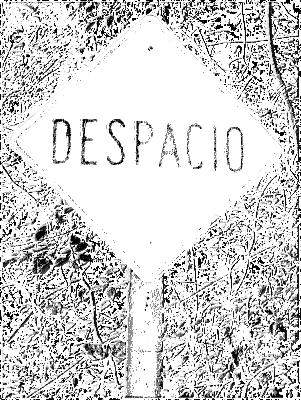} &
 \hspace{-2ex} \includegraphics[scale=0.1]{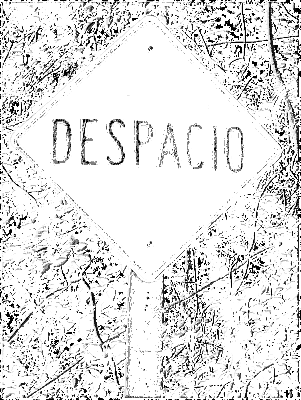} &
 \hspace{-2ex} \includegraphics[scale=0.1]{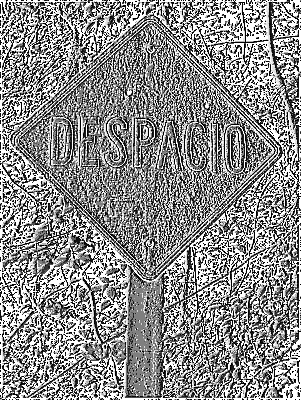} \\
 \hspace{-2ex} \includegraphics[scale=0.1]{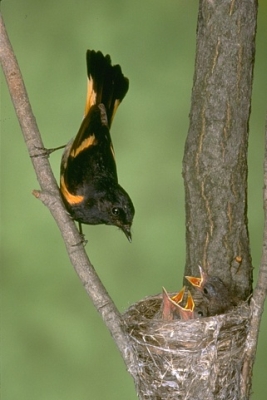}    &
 \hspace{-2ex} \includegraphics[scale=0.1]{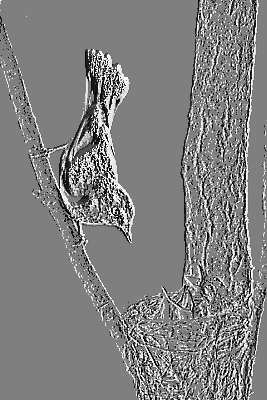} &
 \hspace{-2ex} \includegraphics[scale=0.1]{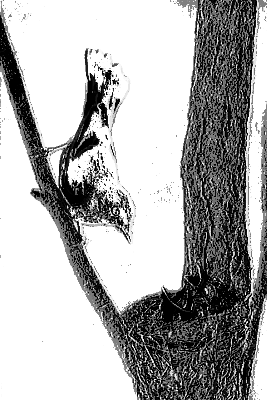} &
 \hspace{-2ex} \includegraphics[scale=0.1]{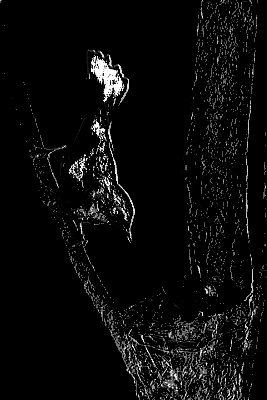} &
 \hspace{-2ex} \includegraphics[scale=0.1]{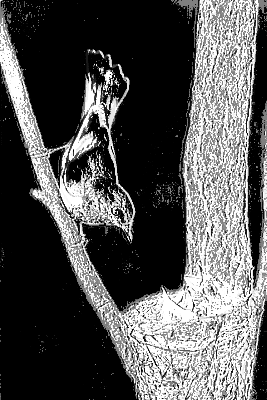} &
 \hspace{-2ex} \includegraphics[scale=0.1]{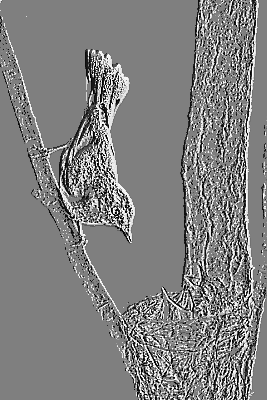} &
 \hspace{-2ex} \includegraphics[scale=0.1]{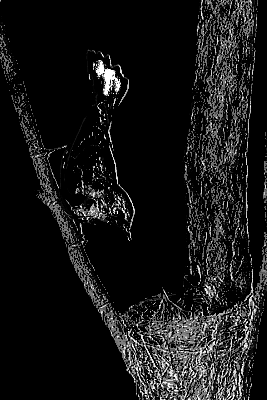} &
 \hspace{-2ex} \includegraphics[scale=0.1]{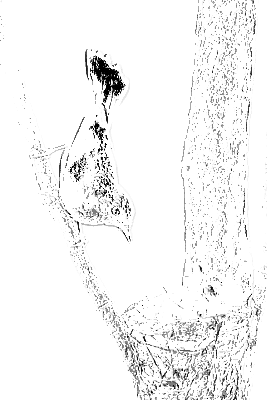} &
 \hspace{-2ex} \includegraphics[scale=0.1]{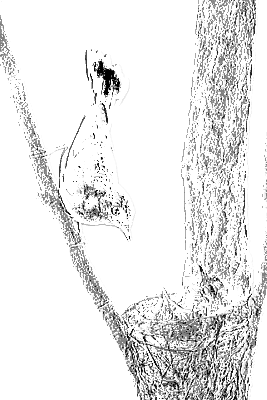} &
 \hspace{-2ex} \includegraphics[scale=0.1]{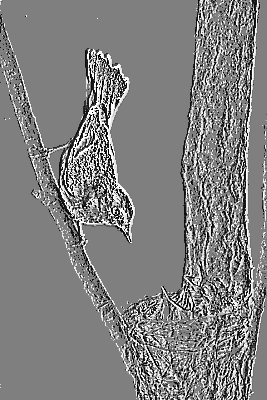} \\
\end{tabular}
\caption{\label{fig:examples_LTP_Pairs_Textures} Micro-texture maps given by LTP on the $9$
 opposing color pairs (for the RGB color space). We can notice that this LTP coding already 
 highlight the salient objects.} 
\end{figure*}

The LTP (Local Ternary Patterns) texture characterization method is
then applied to each opposing color pair to capture the features of
the color micro-textures. At this stage, we obtain 9 grayscale texture
maps which already highlight the salient objects in the image as can
be seen in Figure \ref{fig:examples_LTP_Pairs_Textures}.  We then
consider each texture map as being composed of micro-textures that can
be described by a gray level histogram. As it is not easy to determine
in advance the size of each micro-texture in the image, we chose to
use adaptive windows for each micro-texture. This is why we use
superpixels in our model. To find these superpixels, our model uses
the SLICO (Simple Linear Iterative Clustering with zero parameter)
superpixel algorithm \cite{achanta2012slic} which is a version of SLIC
(Simple Linear Iterative Clustering). The SLICO is a simple, very fast algorithm producing superpixels which has the merit of adhering particularly well to the boundaries (see Figure  \ref{fig:slico_superpixels}) \cite{achanta2012slic}. In Addition, SLICO algorithm (with its default internal parameters), has just one parameter: the number of superpixels desired. Thus, we characterize each pixel of each texture map by the gray level histogram of the superpixel to which it belongs. We thus obtain a histogram map for each texture map. The 9 histogram maps are then concatenated pixel by pixel to have a single histogram map that characterizes the color micro-textures of the image. Each histogram of the latter is then a feature vector for the corresponding pixel.

%
%

The dissimilarity between pixels of the input color image is then
given by the dissimilarity between their feature vectors. We quantify
this dissimilarity thanks to the FastMap method which has the
interesting property of non-linearly reducing in one dimension these
feature vectors while preserving the structure in the data. More
precisely, the FastMap allows to find a configuration, in one
dimension, that preserves, as much as possible, all the (Euclidean)
distance pairs that initially existed between the different (high
dimensional) texture vectors (and that takes into account the
non-linear distribution of the set of feature vectors).  After
normalization between the range $0$ and $1$, the map estimated by the
FastMap produces the Euclidean embedding (in near-linear time) which
can be viewed as a {\em probabilistic} map, {\em i.e.}, with a set of
grey levels with high grayscale values for salient regions and low
values for non-salient areas.

As  Borji {\em and} Itti \cite{borji2012exploiting} stated, almost all
saliency approaches use just one color channel. The latter authors
also argued that employing just one color space does not always lead
to successful outlier detection. Thus, taking into account this
argument, we used, in addition to the RGB color space the color spaces: HSL, LUV and CMY.

Finally, we combine the probabilistic maps obtained from these
color spaces to obtain the desired saliency map. To combine the
probabilistic maps from the different color spaces used, we
reduce for each pixel a vector which is the concatenation of the
averages of the values of the superpixel to which this pixel belongs
successively in all the color spaces used.  In the following section,
we describe the different steps in detail.

%
\begin{figure}[ht]
\centering
\includegraphics[scale=0.28]{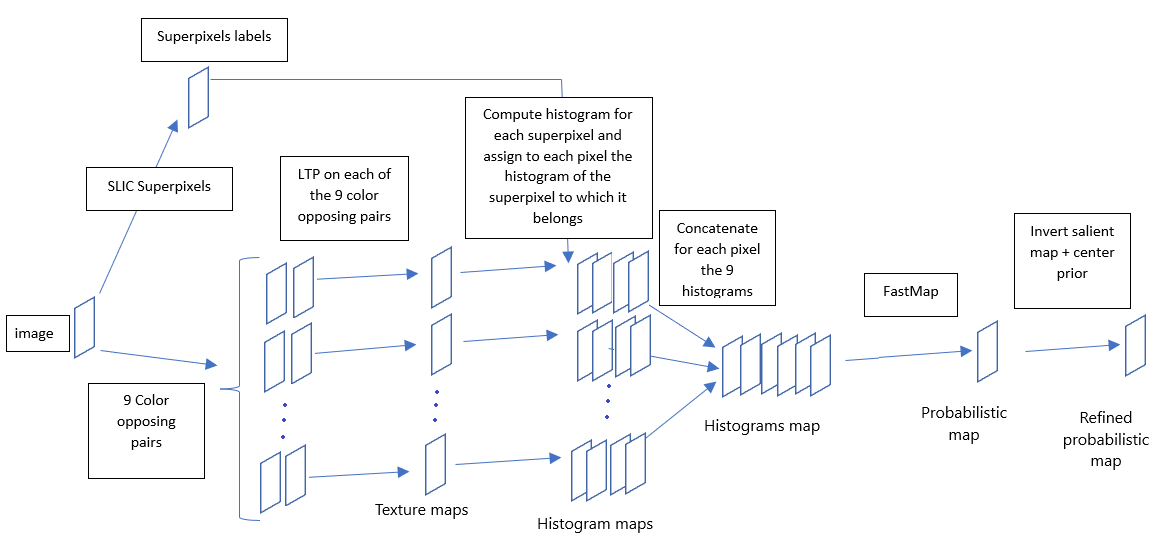}
\vspace{-1ex}
\caption{\label{fig:proposedModel} Proposed model steps to obtain the 
refined probabilistic map from a color space ({\em e.g.} RGB).} 
\end{figure}
%

%
%

\subsection{LTP Texture Characterization on Opposing Color Pairs}
\label{subsec:color_texture}
 
\subsubsection{Local Ternary Patterns (LTP)}

Since LTP ({\em local ternary patterns}) is a kind of generalization
of LBP ({\em local binary patterns}), let's first recall the LBP
technique. 

\begin{figure}[ht]
\centering
\includegraphics{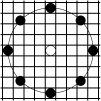}
\caption{\label{fig:LBP_P_R} Example of neighbourhood (black disks) for a pixel 
(central white disk) for $LBP_{P,R}$ code computation : in this case $P = 8$, $R = 4$.} 
\end{figure}

\begin{figure}[ht]
\centering
\begin{tabular}{cccc}
  \includegraphics{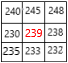} 
& \hspace{-3ex} \includegraphics{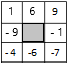} 
& \hspace{-3ex} \includegraphics{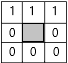} 
& \hspace{-3ex} \includegraphics{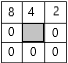}\\
(a) pixel  & (b) after & (c) pattern: & (d)  \\
neighbourhood;  & thresholding; & $00001110$ & code=14 \\
{\scriptsize $g_c=239$}  &  &  &
\end{tabular}
\vspace{0ex}
\caption{\label{fig:LBP_THRESHOLDING} Example of LBP code
  computation for a pixel: LBP code is $2 + 4 + 8 = 14$ in this
  case. }
\end{figure}

The local binary pattern $LBP_{P,R}$ labels each pixel of an image
(see Eq.  \ref{eq:eq_LBP}). The label of a pixel at the position
$(x_c,y_c)$ with  $g_c$ as gray level is a set of P binary digits
obtained by thresholding each gray level value $g_p$ of the $p$
neighbour located at the distance $R$ (see Figure \ref{fig:LBP_P_R}) from
this pixel by the value of the gray level $g_c$ ($p$ is one of the $P$
chosen neighbours). The set of binary digits obtained constitutes the
label of this pixel or its LBP code (see Figure \ref{fig:LBP_THRESHOLDING}).

%
\begin{equation}
\label{eq:eq_LBP}
\mbox{LBP}_{P,R}(x_c,y_c) = \displaystyle\sum_{p=0}^{P-1} s(g_p - g_c )2^p
\end{equation}
%

with $(x_c,y_c)$ being the pixel coordinate and:
%
\[ s(z) =
  \begin{cases}
    1       & \quad \text{if } z \geq 0 \\
    0  & \quad \text{if  } z < 0
  \end{cases}
\]
%
Where $z \! = \! g_p \!-\! g_c$.

Once this code is computed for each pixel, the characterization of
the texture of the image (within a neighbourhood) is approximated by a
discrete distribution (histogram) of LBP codes of $2^P$ bins.
%
%

%
%

The LTP (local ternary patterns) \cite{tan2010enhanced} is an
extension of LBP in which the function $s(z)$ (see
Eq. \ref{eq:eq_LBP}) is defined as follows:
%
\[ s(z) =
  \begin{cases}
    2       & \quad \text{if } z \geq t \\
    1  & \quad \text{if  } |z| < t \\
    0       & \quad \text{if } z \leq -t
  \end{cases}
\]
%
Where $z \!=\! g_p \!-\! g_c$.\\
The basic coding of LTP is, thus expressed as:
%
\begin{equation}
\label{eq:eq_LTP}
\mbox{LTP}_{P,R}(x_c,y_c) = \displaystyle\sum_{p=0}^{P-1} s(g_p - g_c )3^p
\end{equation}
%

\begin{figure}[ht]
\includegraphics[scale=0.3]{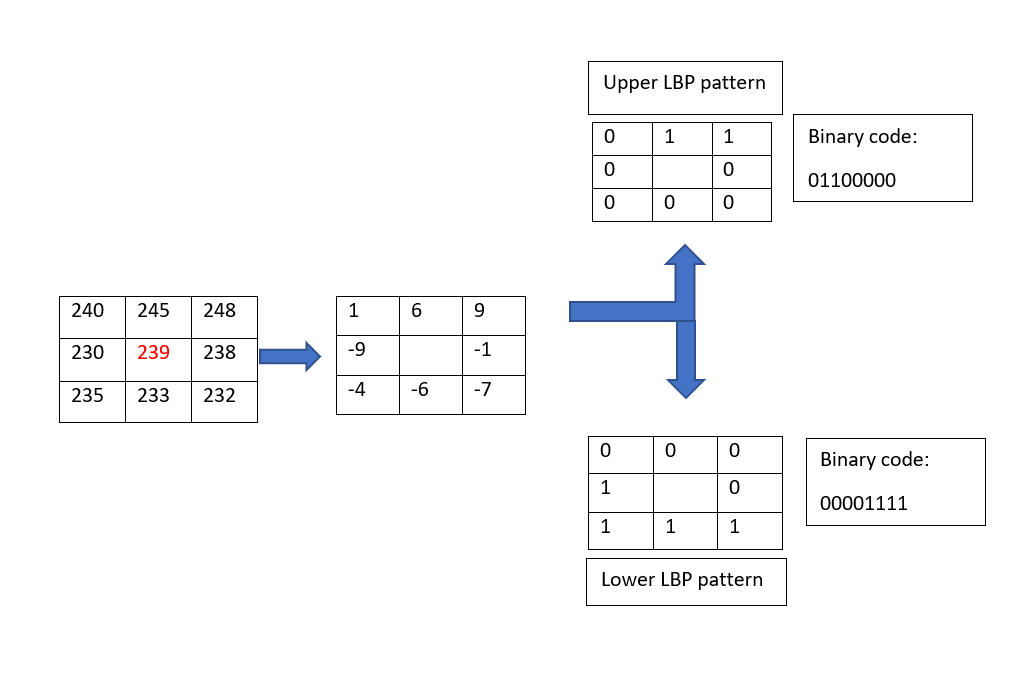} 
\vspace{-2ex}
\caption{\label{fig:LTP_code} Example of LTP code splitting with threshold t=3} 
\end{figure}

Another type of encoding can be obtained by splitting the LTP code
into two codes LBP : Upper LBP code and Lower LBP code (see Figure \ref{fig:LTP_code}). The LTP histogram is then the concatenation
of the histogram of the upper LBP code with that of the lower LBP code
\cite{tan2010enhanced}.\\
In our model we use the LTP basic coding because we use $5$ neighbours
for the central pixel. So the maximum size of the histograms is $3 ^ 5
= 243$. In addition, we requantized the histogram with levels/classes of $75$ bins for first computational reasons (thus greatly reducing the computational time for the next step using the FastMap algorithm while generalizing the feature vector a bit as this operation smoothes the histogram) and we have effectively noticed that this strategy gives slightly better results.

%

\subsubsection{Opposing Color Pairs}

\begin{figure}[ht]
\centering
\includegraphics{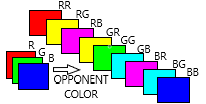}
\vspace{-1ex}
\caption{\label{fig:opponentRGB} Illustration of color opponent on RGB 
(Red Green Blue) color space with its $9$ opposing color pairs ({\em i.e.}, 
RR, RG, RB, GR, GG, GB, BR, BG, BB).
} 
\end{figure}

\begin{figure} [ht]
\centering
\begin{tabular}{cc}
\includegraphics{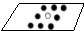} &
\includegraphics{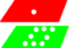}\\
(a) & (b)
\end{tabular}
\vspace{0ex}
\caption{\label{fig:opponentRGB_pixel} (a) Pixel gray LBP code: the
  code for the central pixel ({\em i.e.} white small disk) is computed
  with respect to his neighbours ({\em i.e.} 8 black small disks). (b)
  Pixel opponent color LBP code for RG pair: the central pixel is in
  the first color channel (red) and the neighbours are picked in the
  second channel (green).} 
\end{figure}

To incorporate the color into the texture description, we rely on the
color opponent theory.  We thus used color texture descriptor from
M{\"a}enp{\"a}{\"a} {\em and} Pietik{\"a}inen
\cite{maenpaa2004classification, pietikainen2011computer} called
``Opponent Color LBP''.  This one generalizes the classic LBP, which
normally applies to grayscale textures. So instead of just one LBP
code, one pixel gets a code for every combination of two color
channels ({\em i.e.}, $9$ opposing color pair codes). Example for RGB
channels : RR (Red-Red), RG (Red-Green), RB (Red-Blue), GR
(Green-Red), GG (Green-Green), GB (Green-Blue), BR (Blue-Red), BG
(Blue-Green), BB (Blue-Blue) (see Figure \ref{fig:opponentRGB}). The
central pixel is in the first color channel of the combination and the
neighbours are picked in the second color (see Figure \ref{fig:opponentRGB_pixel} (b)). The histogram that describes
the color micro-texture is the concatenation of the histograms
obtained from each opposing color pair. 

%
%

%
%

\subsection{FastMap : Multi-Dimensional Scaling}
\label{subsec:FastMap}

The FastMap \cite{faloutsos1995FastMap} is an algorithm which
initially was intended to provide a tool allowing to find objects
similar to a given object, to find pairs of the most similar objects
and to visualize distributions of objects in a desired space in order
to be able to identify the main structures in the data, once the
similarity or dissimilarity function is determined. This tool remains
effective even for large collections of data sets, unlike classical
multidimensional scaling (classic MDS).  The FastMap algorithm matches
objects of a certain dimension to points in a $k$-dimensional space
while preserving distances between pairs of objects. This
representation of objects from a large-dimensional space $n$ to a
smaller-dimensional space (dimension $1$ or $2$ or $3$) allows the
visualization of the structures of the distributions in the data or
the acceleration of the search time for queries
\cite{faloutsos1995FastMap}.

As {\em Faloutsos and Lin} \cite{faloutsos1995FastMap} describe it,
the problem solved by FastMap can be represented in two ways. First,
FastMap can be seen as a mean to represent $N$ objects in a
$k$-dimensional space, given the distances between the $N$ objects,
while preserving the distances between pairs of objects. Second, the
FastMap algorithm can also be used in reducing dimensionality while
preserving distances between pairs of vectors. This amounts to
finding, given $N$ vectors having $n$ features each, $N$ vectors in a
space of dimension $k$ - with $n \gg k$ - while preserving the
distances between the pairs of vectors.
To do this, the objects are considered as points in the original
space. The $ \textit {first coordinate axis}$ is the line that
connects the objects called $ \textit {pivots}$. The pivots are chosen
so that the distance separating them is maximum. Thus, to obtain
these pivots, the algorithm follows the steps below:
%
\begin {itemize}
  \item choose arbitrarily an object as the second pivot, {\em i.e.} the object $O_b$;
  \item choose as the first pivot $O_a$, the object furthest from $O_b$ according to the used distance;
  \item replace the second pivot with the furthest object from $O_a$, that is, the object $O_b$;
  \item return the objects $ O_a $ and $ O_b $ as pivots.
\end {itemize}

The axis of the pivots thus constituting the first coordinate axis in
the targeted $k$-dimensional space. All the points representing the
objects are then projected orthogonally on this axis and in the H
hyperplane of $n-1$ dimensions (perpendicular to the first axis
already obtained) connecting the pivot objects $ O_a $ and $ O_b $
along the latter axis. The coordinates of a given object $O_i$ on the
first axis is given by:
%
\begin{equation}
 x_i = \frac{d^2_{a,i} + d^2_{a,b} - d^2_{b,i}}{2 d_{a,b}}
\end{equation}
%
Where $d_{a,i}$, $d_{b,i}$ and $d_{a,b}$ are respectively the distance
between the pivot $O_a$ and object $O_i$, the distance between the
pivot $O_b$ and object $O_i$, the distance between the pivot $O_a$ and
the pivot $O_b$. 
The process is repeated up to the desired dimension, each time expressing:
%
\begin{enumerate}
  \item the new distance $D^{\prime}()$:
    %
    \begin{equation}
    \label{eq:dist_new}
    (D^{\prime} (O^{\prime}_i,O^{\prime}_j))^2 = (D(O_i,O_j))^2 - (x_i - x_j)^2
    \end{equation}
    %
    For simplification, $$D^{\prime} (O^{\prime}_i,O^{\prime}_j) \equiv d^{\prime}_{O^{\prime}_i,O^{\prime}_j }$$
    Where $x_i$ and $x_j$ are the coordinates on the previous axis of respectively the object 
    $O_i$ and $O_j$.
 \item the new pivots $O^{\prime}_a$ and  $O^{\prime}_b$ constituting the new axis,
 \item the coordinate of the projected object  $O^{\prime}_i$ on the new axis:
   %
   \begin{equation}
    \label{eq:dist_between_O_prime_a_and_O_prime_i}
     x^{\prime}_i = \frac{d^{\prime 2}_{a^{\prime},i} + d^{\prime 2}_{a^{\prime},b^{\prime}} 
                    - d^{\prime 2}_{b^{\prime},i}}{2 d^{\prime}_{a^{\prime},b^{\prime}}}
    \end{equation}
  %
\end{enumerate}

$O_{a^\prime}$ and $O_{b^\prime}$ are the new pivots according to the
new distance expression $D^\prime()$. The line that connects them is
therefore the new axis.

\vspace{1ex}
After normalization between the range $0$ and $1$, the map estimated
by the FastMap generates a {\em probabilistic} map, {\em i.e.}, with a
set of grey levels with high grayscale values for salient regions and
low values for non-salient areas. Nevertheless, in some (rare)
cases, the map estimated by the FastMap algorithm, can possibly
present a set of grey levels whose amplitude values would be in the
complete opposite direction ({\em i.e.}, low grayscale values for
salient regions and high values for non-salient areas). In order to
put this grayscale mapping in the right direction (with high grayscale
values associated with salient objects), we simply use the fact that a
salient object/region is more likely to appear in the center of the
image (or conversely unlikely on the edges of the image). To this
end, we compute the Pearson correlation coefficient between the
saliency map obtained by the FastMap and a rectangle, with maximum
intensity value and about half the size of the image, and located in
the center of the image. If the correlation coefficient is negative
(anti-correlation), we invert the signal ({\em i.e.}, associate  to
each pixel its complementary gray value).

%
\begin{figure*}[ht]
\centering
\begin{tabular}{ccccccccccc}
NUMBER & 1 & 2 & 3 & 4 & 5 & 6 & 7 & 8 & 9 & 10\\
IMAGE & 
\hspace{-2ex} \includegraphics[scale=0.1]{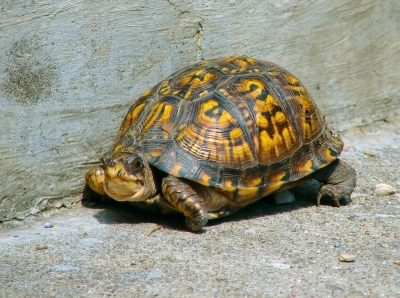} &
\hspace{-2ex} \includegraphics[scale=0.1]{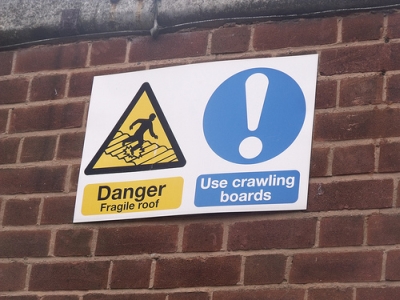} &
\hspace{-2ex} \includegraphics[scale=0.1]{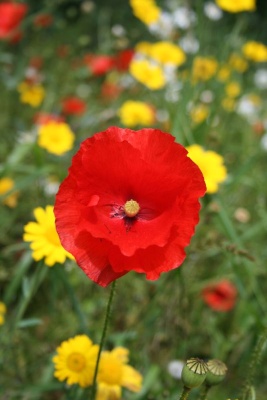} &
\hspace{-2ex} \includegraphics[scale=0.1]{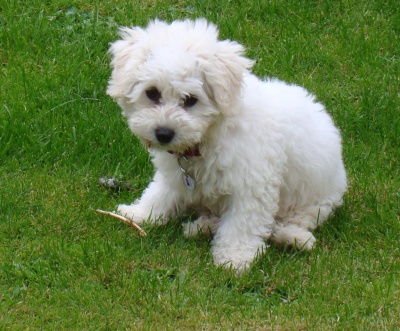} &
\hspace{-2ex} \includegraphics[scale=0.1]{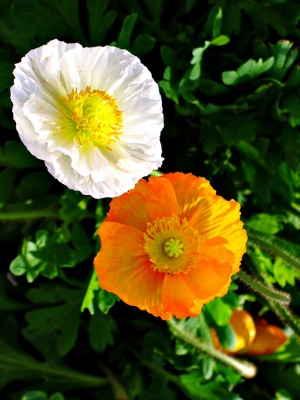} &
\hspace{-2ex} \includegraphics[scale=0.1]{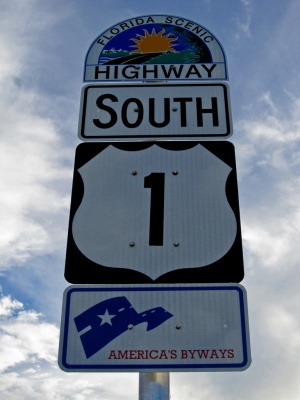} &
\hspace{-2ex} \includegraphics[scale=0.1]{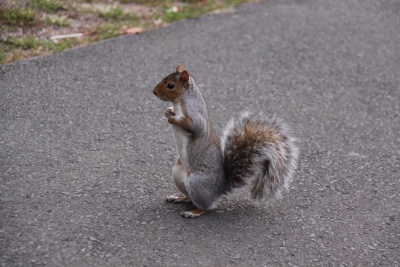} &
\hspace{-2ex} \includegraphics[scale=0.1]{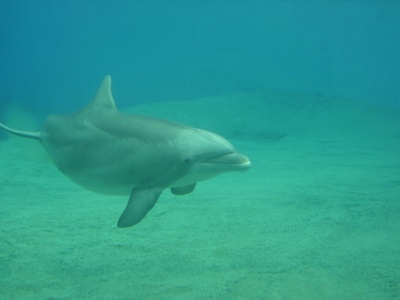} &
\hspace{-2ex} \includegraphics[scale=0.1]{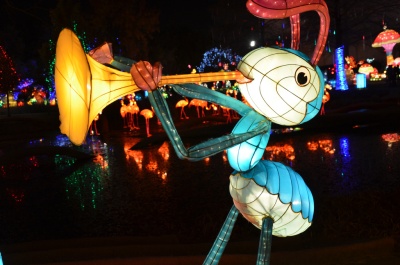} &
\hspace{-2ex} \includegraphics[scale=0.1]{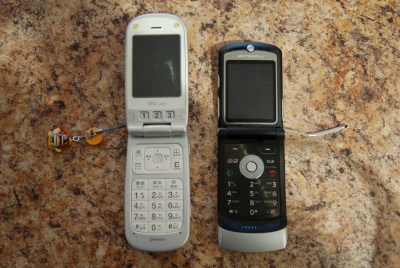} \\
GT &       
\hspace{-2ex} \includegraphics[scale=0.1]{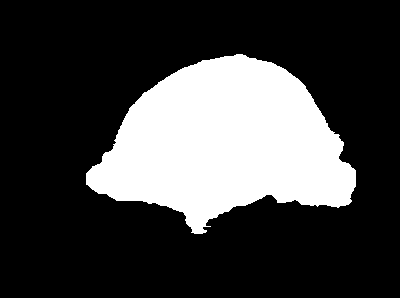} &
\hspace{-2ex} \includegraphics[scale=0.1]{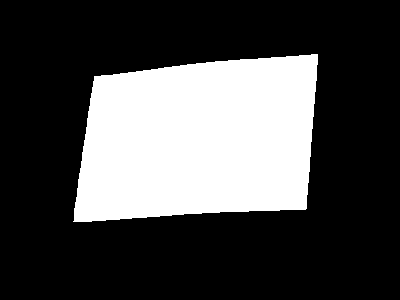} &
\hspace{-2ex} \includegraphics[scale=0.1]{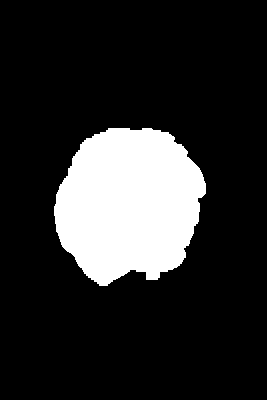} &
\hspace{-2ex} \includegraphics[scale=0.1]{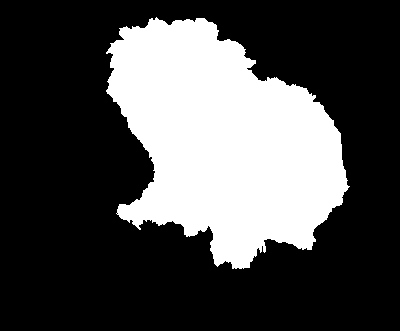} &
\hspace{-2ex} \includegraphics[scale=0.1]{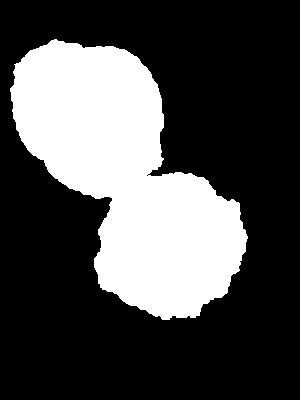} &
\hspace{-2ex} \includegraphics[scale=0.1]{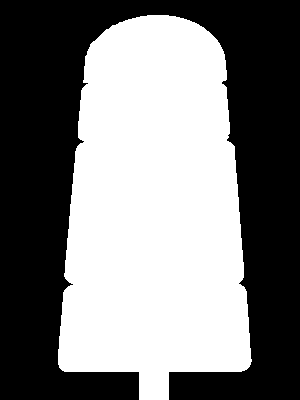} &
\hspace{-2ex} \includegraphics[scale=0.1]{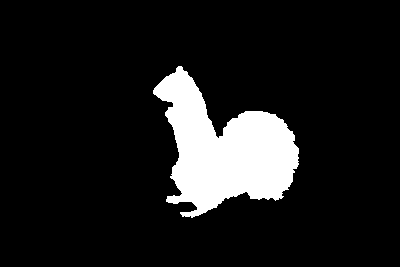} &
\hspace{-2ex} \includegraphics[scale=0.1]{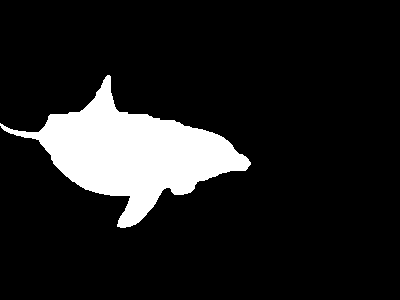} &
\hspace{-2ex} \includegraphics[scale=0.1]{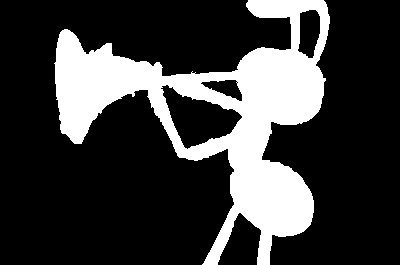} &
\hspace{-2ex} \includegraphics[scale=0.1]{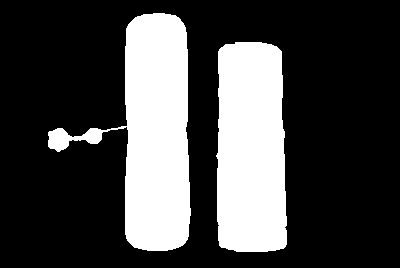} \\
CHS\cite{shi2016hierarchical} &       
\hspace{-2ex} \includegraphics[scale=0.1]{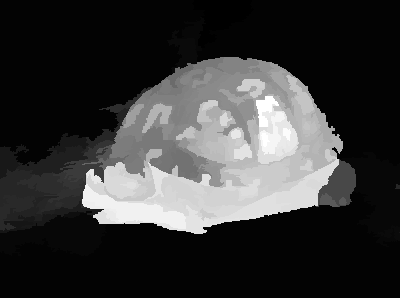} &
\hspace{-2ex} \includegraphics[scale=0.1]{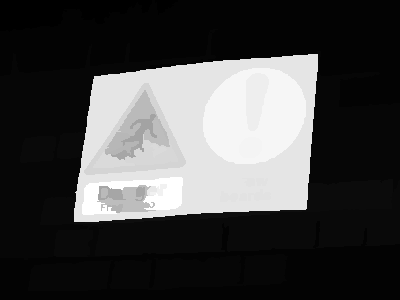} &
\hspace{-2ex} \includegraphics[scale=0.1]{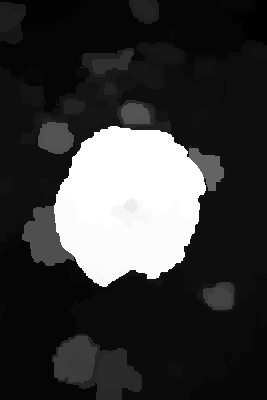} &
\hspace{-2ex} \includegraphics[scale=0.1]{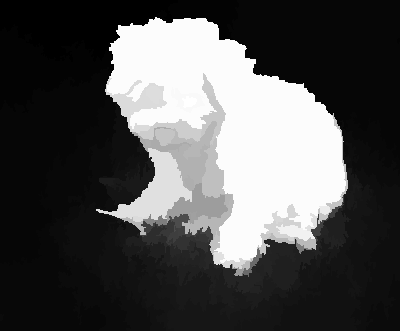} &
\hspace{-2ex} \includegraphics[scale=0.1]{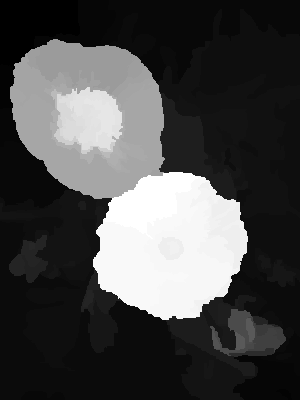} &
\hspace{-2ex} \includegraphics[scale=0.1]{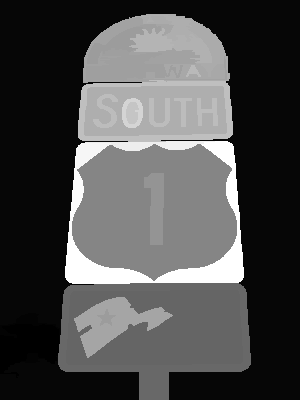} &
\hspace{-2ex} \includegraphics[scale=0.1]{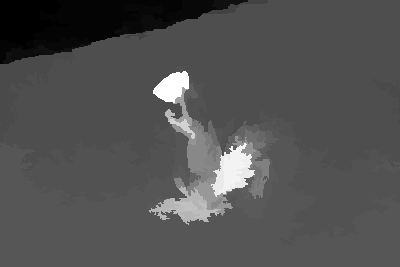} &
\hspace{-2ex} \includegraphics[scale=0.1]{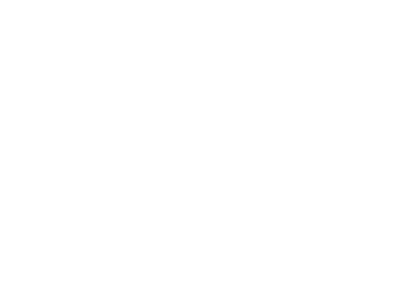} &
\hspace{-2ex} \includegraphics[scale=0.1]{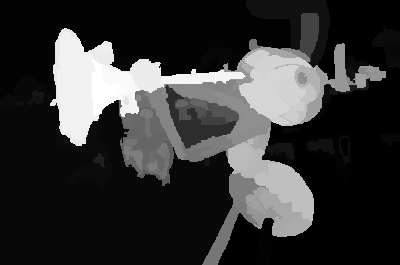} &
\hspace{-2ex} \includegraphics[scale=0.1]{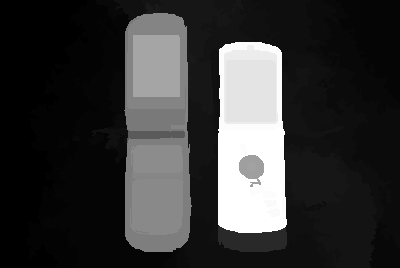} \\
HS\cite{yan2013hierarchical} &       
\hspace{-2ex} \includegraphics[scale=0.1]{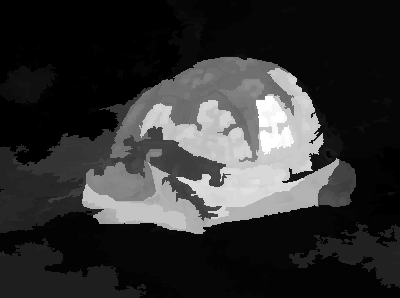} &
\hspace{-2ex} \includegraphics[scale=0.1]{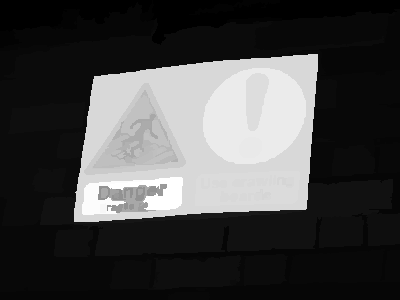} &
\hspace{-2ex} \includegraphics[scale=0.1]{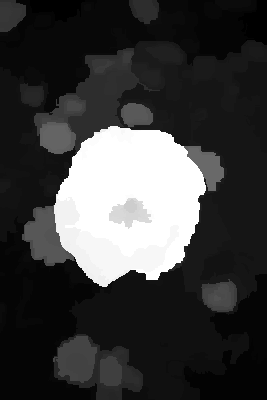} &
\hspace{-2ex} \includegraphics[scale=0.1]{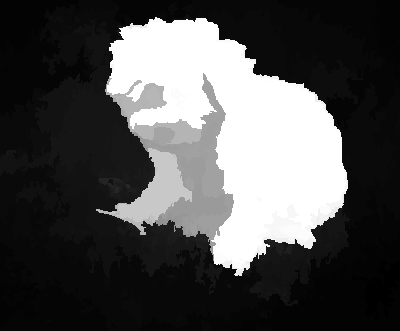} &
\hspace{-2ex} \includegraphics[scale=0.1]{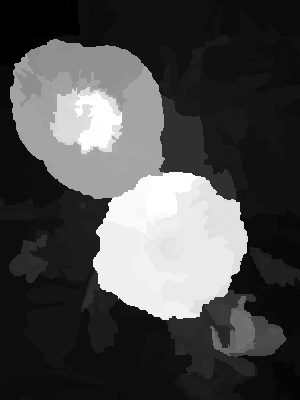} &
\hspace{-2ex} \includegraphics[scale=0.1]{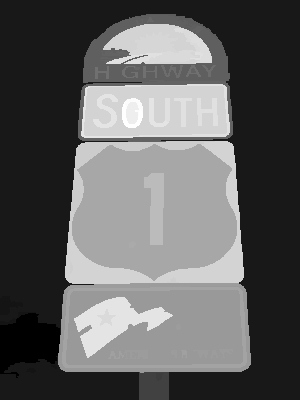} &
\hspace{-2ex} \includegraphics[scale=0.1]{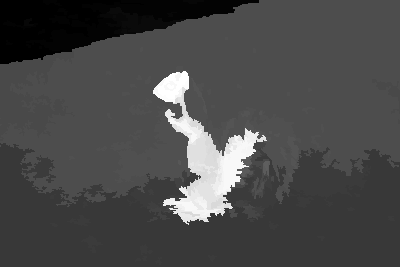} &
\hspace{-2ex} \includegraphics[scale=0.1]{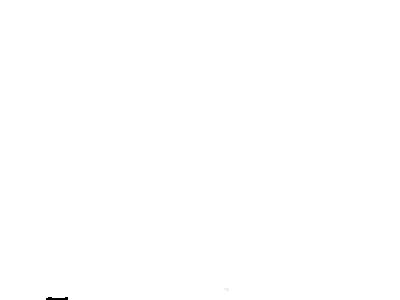} &
\hspace{-2ex} \includegraphics[scale=0.1]{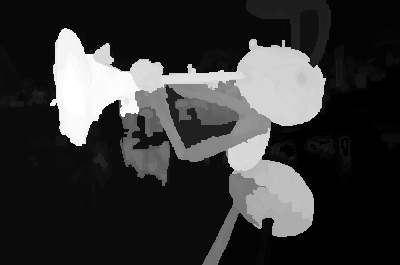} &
\hspace{-2ex} \includegraphics[scale=0.1]{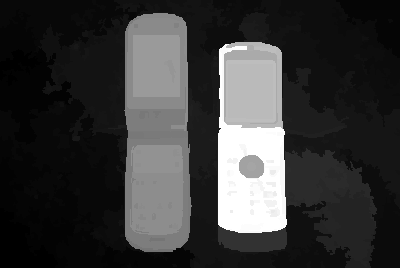} \\
OURS &       
\hspace{-2ex} \includegraphics[scale=0.1]{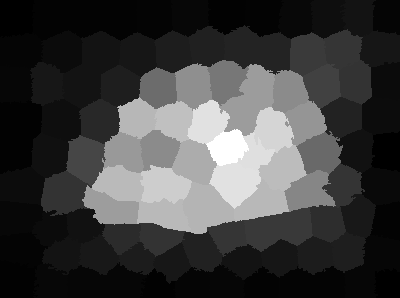} &
\hspace{-2ex} \includegraphics[scale=0.1]{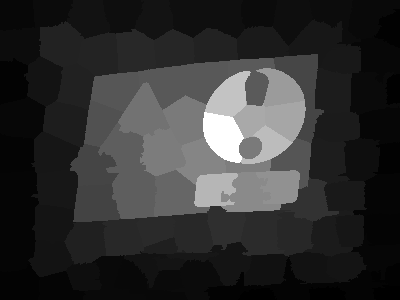} &
\hspace{-2ex} \includegraphics[scale=0.1]{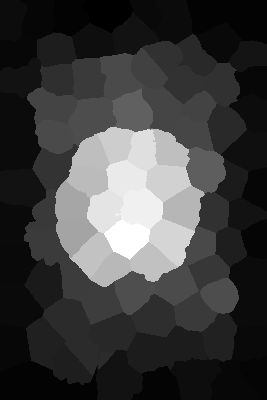} &
\hspace{-2ex} \includegraphics[scale=0.1]{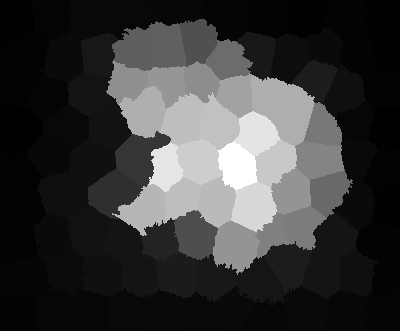} &
\hspace{-2ex} \includegraphics[scale=0.1]{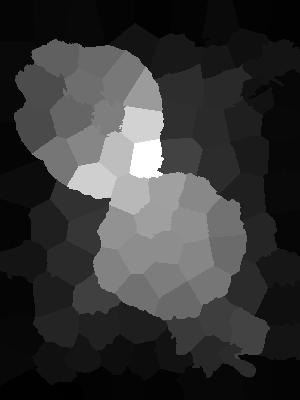} &
\hspace{-2ex} \includegraphics[scale=0.1]{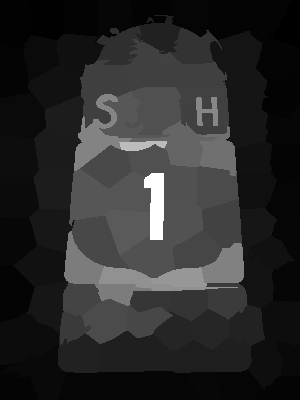} &
\hspace{-2ex} \includegraphics[scale=0.1]{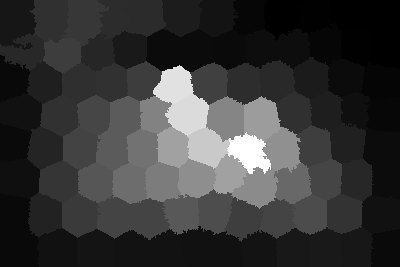} &
\hspace{-2ex} \includegraphics[scale=0.1]{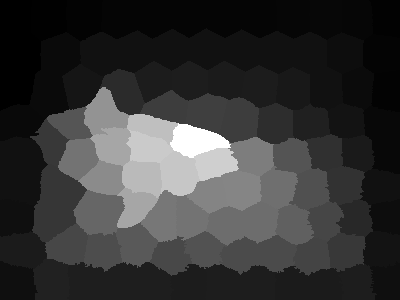} &
\hspace{-2ex} \includegraphics[scale=0.1]{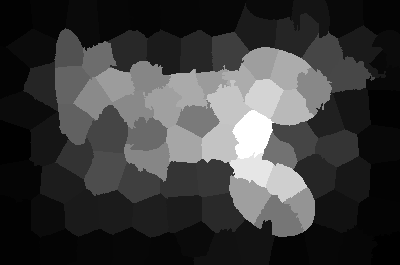} &
\hspace{-2ex} \includegraphics[scale=0.1]{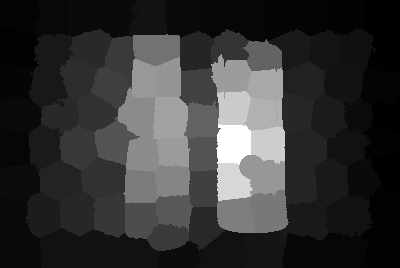} \\    
\end{tabular}
\caption{\label{fig:IMAGES_COMPARISON_HS_CHS_OURS} Comparison of some result images 
for HS\cite{yan2013hierarchical}, CHS\cite{shi2016hierarchical} and our model. For 
image number 8, the HS\cite{yan2013hierarchical} and CHS\cite{shi2016hierarchical} 
models find white salient maps (GT: Ground Truth).} 
\end{figure*}

%

\vspace{0ex}
\section{Experimental Results}
\label{ExperimentalResults}

In this section, we present our salient objects detection model's
results.  In order to obtain the LTP$_{P,R}$  pixel's code (LTP code
for simplification), we used an adaptive threshold. Let a pixel at
position $(x_c, y_c)$ with value $g_c$, the threshold for its LTP code
is the tenth of the pixel's value: $t = \frac{g_c}{10}$ (see
Eq. \ref{eq:eq_LTP}). We chose this threshold because empirically it
is this value that has given better results.  The number of neighbours
P around the pixel on a radius R used to find its  LTP code in our
model is $P=5$ and $R=1$. Thus the maximum value of the  LTP code in
our case is $3^5 - 1 = 242$. This makes the maximum size of the histogram characterizing the micro-texture in an opposing color pair
to be $3^5 = 243$ which is then requantized with levels/classes of 75 bins (see Section \ref{subsec:color_texture}). The superpixels that we use as adaptive windows to characterize the color micro-textures are obtained thanks to SLICO (Simple Linear Iterative Clustering with zero
parameter) algorithm which is faster and exhibits state-of-the-art
boundary adherence. Its only parameter is the number of superpixels
desired and is set to $100$ in our model (which is also the value
recommended by the author of the SLICO algorithm).  Finally, we use in
the combination to obtain the final saliency map the color spaces RGB,
HSL, LUV and CMY.
We chose, for our experiments, images from public datasets, the most
widely used in the salient objects detection field
\cite{borji2015salient} such as Extended Complex Scene Saliency
Dataset (ECSSD) and Microsoft Research Asia 10,000 (MSRA10K). The
ECSSD contains $1000$ natural images and their ground truth. Many of
its images are semantically meaningful, but structurally complex for
saliency detection \cite{shi2016hierarchical}. The MSRA10K contains
$10 \, 000$ images and $10 \, 000$ manually obtained binary saliency
maps corresponding to their ground truth \cite{cheng2015global,
  borji2015salient}.

\vspace{2ex}
\begin{table}[ht]
\caption{ Our model's $F_{\beta}$ measure and MAE results 
for ECSSD and MSRA10K datasets.} 
\label{tab:taux_FM_MAE}
\centering
\begin{tabular}{|c|c|c|}
 \hline
 & ECSSD &  MSRA10K \\
 \hline
 $F_{\beta}$ measure & $0.729$  &  $0.781$ \\
 \hline
 MAE & $0.257$ & $0.222$ \\
 \hline
\end{tabular}
\vspace{1ex}
\end{table}
%
\begin{table}[ht]
\caption{ Our model's
  $F_{\beta}$ measure and MAE results compared with some
  state-of-the-art models from Borji { \em et al.}
  \cite{borji2015salient} for ECSSD dataset (for MAE, the smaller
  value is the best).}
  \label{tab:taux_FM_MAE_STATE_ART_OURS}
\centering
\begin{tabular}{|c|c|c|c|c|c|}
 \hline
 & AC \cite{achanta2008salient} &  CA\cite{goferman2012context}  
 & HC\cite{cheng2015global}     & HS\cite{yan2013hierarchical} 
 & OURS \\
 \hline
 MAE & $0.265$  &  $0.310$  &  $0.331$  &  $0.228$ &  $0.257$\\
 \hline
 $F_{\beta}$  & $0.411$  &  $0.515$  &  $0.460$  &  $0.731$ &  $0.729$\\ 
 \hline
\end{tabular}
\vspace{1ex}
 
\end{table}

\vspace{1ex}
Our saliency maps are of good quality (see Figure
 \ref{fig:IMAGES_COMPARISON_HS_CHS_OURS}).) as shown by the visual
comparison with some of them and two state-of-the-art models
(``Hierarchical saliency detection'': HS\cite{yan2013hierarchical} and ``Hierarchical image saliency detection on extended CSSD'': CHS\cite{shi2016hierarchical}
models).  We used for evaluation of our salient objects detection
model the Mean Absolute Error (MAE), the Precision-Recall curve (PR),
$F_{\beta}$ measure curve and the $F_{\beta}$ measure with $\beta^2 =
0.3$. Table \ref{tab:taux_FM_MAE} shows the  $F_{\beta}$ measure and
the Mean Absolute Error (MAE) of our model on ECSSD and MSRA10K
datasets.

\begin{figure*}[ht]
\centering
\includegraphics[scale=0.47]{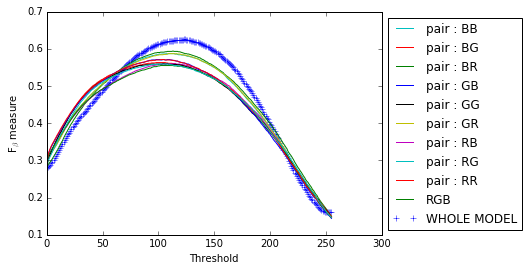}\\
\vspace{-1ex}
\caption{\label{fig:F_beta_measure_PAIRS_RGB_SPACE_WHOLE_MODEL}  $F_{\beta}$ measure curves 
for opposing color pairs, RGB color space and the whole model on ECSSD dataset.} 
\end{figure*}

\begin{figure}[!b]
\vspace {-4mm}
\centering
\includegraphics[scale=0.47]{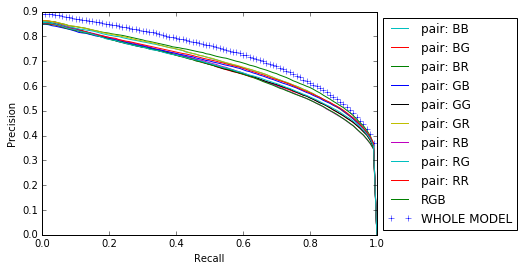}\\
\vspace{-1ex}
\caption{\label{fig:PR_PAIRS_RGB_SPACE_WHOLE_MODEL} Precision-Recall curves for opposing color 
pairs, RGB color space and the whole model on ECSSD dataset.} 
\end{figure}

\begin{figure}[!b]
\vspace {-4mm}
\centering
\includegraphics[scale=0.47]{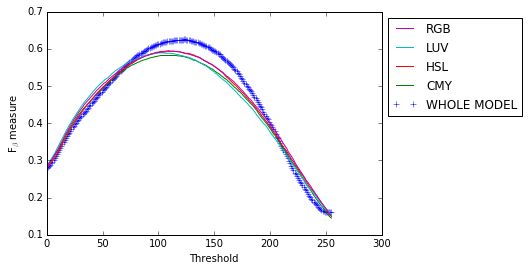}\\
\vspace{-1ex}
\caption{\label{fig:F_beta_measure_COLOR_SPACES_WHOLE_MODEL} $F_{\beta}$ measure 
curves for color spaces RGB, HSL, LUV and CMY and the whole model on  ECSSD dataset.} 
\end{figure}

\begin{figure}[!b]
\vspace {-4mm}
\centering
\includegraphics[scale=0.5]{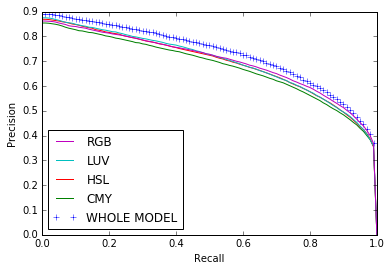}\\
\vspace{-1ex}
\caption{\label{fig:PR_COLOR_SPACES_WHOLE_MODEL} Precision-Recall curves  for color spaces 
RGB, HSL, LUV and CMY and the whole model on  ECSSD dataset.} 
\end{figure}

\vspace{1ex}
Our results also show that combining the opposing color pairs improves
the individual contribution of each pair to the F$_{\beta}$ measure and the Precision-Recall as shown for the RGB color space by the F$_{\beta}$ measure curve (Figure \ref{fig:F_beta_measure_PAIRS_RGB_SPACE_WHOLE_MODEL}) and the Precision-Recall curve (Figure \ref{fig:PR_PAIRS_RGB_SPACE_WHOLE_MODEL}).  The combination of the color spaces RGB, HSL, LUV and CMY improves also the final result as it can be seen on the $F_{\beta}$ measure curve and the precision-recall curve (see Figure
\ref{fig:F_beta_measure_COLOR_SPACES_WHOLE_MODEL} and Figure
\ref{fig:PR_COLOR_SPACES_WHOLE_MODEL}). 

\begin{figure}[!b]
\vspace {-4mm}
\centering
\includegraphics[scale=0.5]{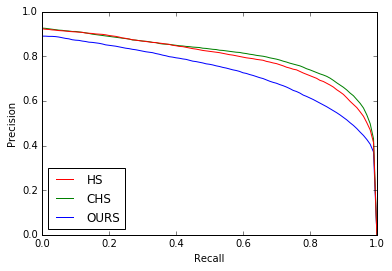}\\
\vspace{-1ex}
\caption{\label{fig:PR_HS_CHS_OURS} Precision-Recall curves for HS\cite{yan2013hierarchical}, 
CHS \cite{shi2016hierarchical} models and ours on ECSSD dataset.} 
\end{figure}

\begin{figure}[!b]
\vspace {-4mm}
\centering
\includegraphics[scale=0.5]{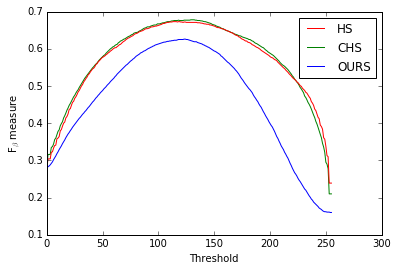}\\
\vspace{-1ex}
\caption{\label{fig:F_beta_measure_HS_CHS_OURS}  $F_{\beta}$ measure curves 
for HS \cite{yan2013hierarchical}, CHS \cite{shi2016hierarchical} models and 
ours on  ECSSD dataset.} 
\end{figure}

\vspace{1ex}
We compared the MAE (Mean Absolute Error) and F$_{\beta}$ measure of
our model with the $29$ state-of-the-art models from Borji {\em et
  al.}\cite{borji2015salient} and our model outperformed $11$
models. Table \ref{tab:taux_FM_MAE_STATE_ART_OURS} shows  the MAE
(Mean Absolute Error) and F$_{\beta}$ measures values for the ECSSD
dataset of some models.  Finally, we compared our model with the two
state-of-the-art HS\cite{yan2013hierarchical} and CHS
\cite{shi2016hierarchical} models with respect to the precision-recall  and F$_{\beta}$ measure curves. We see that our model still
performs well (see Figure \ref{fig:PR_HS_CHS_OURS} and Figure \ref{fig:F_beta_measure_HS_CHS_OURS}).

%
%
%
%
%
%
%
%
%
%
%
%

\vspace{0ex}
\section{Discussion}
\label{Discussion}

\begin{figure}[!b]
\vspace {-4mm}
\centering
\includegraphics[scale=0.5]{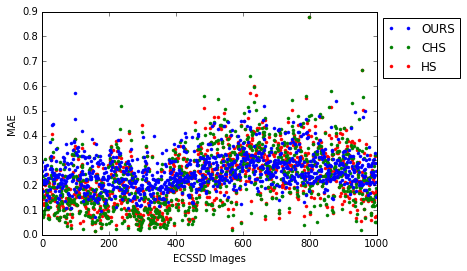}\\
\vspace{-1ex}
\caption{\label{fig:dispersion_mae_ours_hs_chs} Comparison of the MAE measure dispersion for our model and the HS \cite{yan2013hierarchical}, CHS \cite{shi2016hierarchical} models on ECSSD dataset (for MAE, the smaller value is the best).}
\end{figure}

  Our model has less dispersed MAE measures than the HS\cite{yan2013hierarchical} and CHS\cite{shi2016hierarchical} models  which are among the best models of the state-of-the-art. This can be observed in Figure \ref{fig:dispersion_mae_ours_hs_chs} but also shown by the standard deviation which for our model is 0.071 (mean = 0.257), for HS\cite{yan2013hierarchical} is 0.108 (mean = 0.227), for CHS\cite{shi2016hierarchical} is 0.117 (mean = 0.226). For HS\cite{yan2013hierarchical} the relative error between the two standard deviations is $\frac{(0.108-0.071) \times 100}{0.071}= 52.11\%$ while for CHS\cite{shi2016hierarchical} is $\frac{(0.117-0.071) \times 100}{0.071}= 64.78\%$.
%
%
%
%

\begin{figure}[!b]
\vspace {-4mm}
\centering
\includegraphics[scale=0.5]{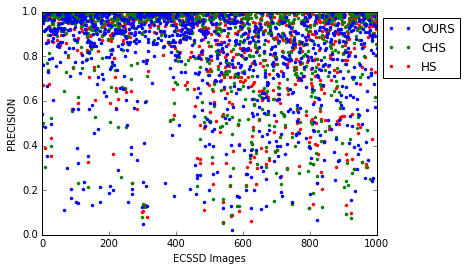}\\
\vspace{-1ex}
\caption{\label{fig:dispersion_precision_ours_hs_chs}  Comparison of the precision measure dispersion for our model and the HS \cite{yan2013hierarchical}, CHS \cite{shi2016hierarchical} models on ECSSD dataset.}
\end{figure}

  Our model is stable on new data. Indeed, a model with very few internal parameters is supposed to be more stable for different datasets. Also, we noticed that nearly 500 first image numbers of the ECSSD dataset are less complex than the rest of the images in this dataset by observing the different measures (see Table \ref{tab:taux_drop_measure_0_500_1000} and Figure \ref{fig:dispersion_mae_ours_hs_chs} and Figure \ref{fig:dispersion_precision_ours_hs_chs}). But it is clear that the drop in performance over the last 500 images from the ECSSD dataset is less pronounced for our model than for the HS \cite{yan2013hierarchical} and CHS \cite{shi2016hierarchical} models (see Table \ref{tab:taux_drop_measure_0_500_1000}). This can be explained by the stability of our model (we used to compute these measures except for MAE a threshold, for each image, which gives the best $F_\beta$ measure. It should also be noted that the images are ordered only by their numbers in the ECSSD dataset).

\vspace{2ex}
\begin{table*}[ht]

\caption{  Performance drop for Precision and MAE measures with respect to image numbers 0 to 500(*) and 500 to 1000(**) of the ECSSD dataset (for MAE, the smaller value is the best).}
\label{tab:taux_drop_measure_0_500_1000}
\centering
\begin{tabular}{|c|c|c|c||c|c|c|}
 \hline
  &\multicolumn{3}{|c||}{Precision}&\multicolumn{3}{|c|}{MAE}\\ \cline{2-7}
                      & ours  &   HS  & CHS   & ours  & HS    & CHS  \\
 \hline 
 (*)    & 0.832 & 0.919 & 0.921 & 0.234 & 0.176 & 0.172\\
 \hline 
(**) & 0.737 & 0.791 & 0.791 & 0.279 & 0.278 & 0.280\\
 \hline 
 \textbf{Gap} & \textbf{0.095} & \textbf{0.128} & \textbf{0.130} & \textbf{0.045} & \textbf{0.102} & \textbf{0.108}  \\ 
 \hline
\end{tabular}
\vspace{1ex}

\end{table*}
%

%
%
%

\begin{figure}[!b]
\vspace {-4mm}
\centering
\includegraphics[scale=0.5]{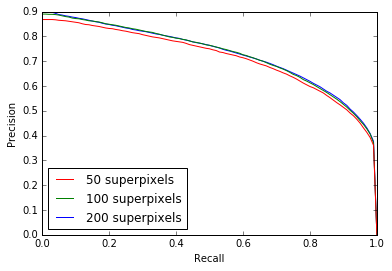}\\
\vspace{-1ex}
\caption{\label{fig:PR_50_100_200_SPX}  Precision-Recall model's curves for 50, 100, 200 superpixels (ECSSD dataset).}
\end{figure}

\begin{figure}[!b]
\vspace {-4mm}
\centering
\includegraphics[scale=0.5]{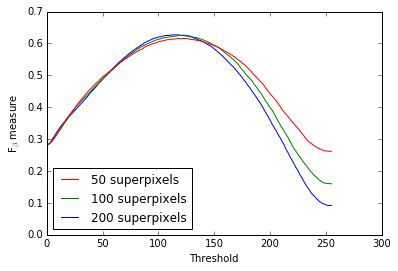}\\
\vspace{-1ex}
\caption{ $F_{\beta}$ measure model's curves for 50, 100, 200 superpixels (ECSSD dataset).}
\label{fig:F_beta_measure_50_100_200_SPX}
\end{figure}

 Our model is also relatively stable for an increase or decrease of its unique internal parameter. Indeed, by increasing or decreasing the number of superpixels, which is the only parameter of the SLICO algorithm, we find that there is almost no change in the results as shown by the MAE and $F_{\beta}$ measure (see Table \ref{tab:taux_FM_MAE_50_100_200}) and $F_{\beta}$ measure and precision-recall curves for 50, 100 and 200 superpixels (see Figure \ref{fig:PR_50_100_200_SPX} and Figure \ref{fig:F_beta_measure_50_100_200_SPX}).

\vspace{2ex}
\begin{table}[ht]
\caption{ Our model's $F_{\beta}$ measure and MAE results for 50, 100 and 200 superpixels (ECSSD dataset).}
\label{tab:taux_FM_MAE_50_100_200}
\centering
\begin{tabular}{|c|c|c|c|}
 \hline
 \textbf{Superpixels} & \textbf{50}  &  \textbf{100} & \textbf{200} \\
 \hline
\textbf{ $F_{\beta}$ measure} & $0.722$  &  $0.729$ &  $0.725$ \\
 \hline
 \textbf{MAE} & $0.257$ & $0.257$ & $0.257$ \\
 \hline
\end{tabular}
\vspace{1ex}

\end{table}
%

%
%
%
%
%


\section{Conclusions}\label{sec:conclusions}

In this work, we presented a simple nearly parameter-free model  for
the estimation of saliency maps. We tested our model on the complex
ECSSD dataset and on the MSRA10K dataset on which the average measures
of MAE = $0.257$ and F$_{\beta}$ measure = $0.729$.\\
The novelty of our model is that it only uses the textural feature
after incorporating the color information into  these textural
features thanks to the opposing color pairs theory of a given color
space. This is made possible by the LTP (Local Ternary Patterns)
texture descriptor which, being an extension of LBP (Local Binary
Patterns), inherits its strengths while being less sensitive to noise
in uniform regions. Thus, we characterize each pixel of the image by a
feature vector given by a color micro-texture obtained thanks to SLICO
superpixel algorithm. In addition, the FastMap algorithm reduces each
of these feature vectors to one dimension while taking into account
the non-linearities of these vectors and preserving their
distances. This means that our saliency map combines local and global
approaches in a single approach and does so in almost linear
complexity times.\\
 In our model, we used RGB, HSL, LUV and CMY color spaces. Our model is therefore perfectible if we increase the number of color spaces (uncorrelated) to be merged.\\
As shown by the results we obtained, this strategy generates a model
which is very promising, since it is quite different from existing
saliency detection methods using the classical color contrast strategy
between a region and the other regions of the image and consequently
it could thus be efficiently combined with these methods for
better performance.  In addition, it should be noted that  this
strategy of integrating color into local textural patterns could also
be interesting to study with  deep learning techniques or
convolutional neural networks (CNNs) to further improve the quality
of saliency maps.


\bibliographystyle{unsrt}
\bibliography{Article_Salient_Maps_LTP}

\begin{thebibliography}{10}

\bibitem{parkhurst2002modeling}
Derrick Parkhurst, Klinton Law, and Ernst Niebur.
\newblock Modeling the role of salience in the allocation of overt visual
  attention.
\newblock {\em Vision research}, 42(1):107--123, 2002.

\bibitem{itti2005models}
Laurent Itti.
\newblock Models of bottom-up attention and saliency.
\newblock In {\em Neurobiology of attention}, pages 576--582. Elsevier, 2005.

\bibitem{itti2001computational}
Laurent Itti and Christof Koch.
\newblock Computational modelling of visual attention.
\newblock {\em Nature reviews neuroscience}, 2(3):194--203, 2001.

\bibitem{baluch2011mechanisms}
Farhan Baluch and Laurent Itti.
\newblock Mechanisms of top-down attention.
\newblock {\em Trends in neurosciences}, 34(4):210--224, 2011.

\bibitem{treisman1988features}
Anne Treisman.
\newblock Features and objects: The fourteenth bartlett memorial lecture.
\newblock {\em The quarterly journal of experimental psychology},
  40(2):201--237, 1988.

\bibitem{wolfe1989guided}
Jeremy~M Wolfe, Kyle~R Cave, and Susan~L Franzel.
\newblock Guided search: an alternative to the feature integration model for
  visual search.
\newblock {\em Journal of Experimental Psychology: Human perception and
  performance}, 15(3):419, 1989.

\bibitem{koch1987shifts}
Christof Koch and Shimon Ullman.
\newblock Shifts in selective visual attention: towards the underlying neural
  circuitry.
\newblock In {\em Matters of intelligence}, pages 115--141. Springer, 1987.

\bibitem{borji2019salient}
Ali Borji, Ming-Ming Cheng, Qibin Hou, Huaizu Jiang, and Jia Li.
\newblock Salient object detection: A survey.
\newblock {\em Computational visual media}, pages 1--34.

\bibitem{borji2012state}
Ali Borji and Laurent Itti.
\newblock State-of-the-art in visual attention modeling.
\newblock {\em IEEE transactions on pattern analysis and machine intelligence},
  35(1):185--207, 2012.

\bibitem{itti2004automatic}
Laurent Itti.
\newblock Automatic foveation for video compression using a neurobiological
  model of visual attention.
\newblock {\em IEEE transactions on image processing}, 13(10):1304--1318, 2004.

\bibitem{li2020saliency}
Jinjiang Li, Xiaomei Feng, and Hui Fan.
\newblock Saliency-based image correction for colorblind patients.
\newblock {\em Computational Visual Media}, 6(2):169--189, 2020.

\bibitem{gao2015database}
Yuan Gao, Miaojing Shi, Dacheng Tao, and Chao Xu.
\newblock Database saliency for fast image retrieval.
\newblock {\em IEEE Transactions on Multimedia}, 17(3):359--369, 2015.

\bibitem{pieters2004attention}
Rik Pieters and Michel Wedel.
\newblock Attention capture and transfer in advertising: Brand, pictorial, and
  text-size effects.
\newblock {\em Journal of Marketing}, 68(2):36--50, 2004.

\bibitem{wong2009saliency}
Lai-Kuan Wong and Kok-Lim Low.
\newblock Saliency-enhanced image aesthetics class prediction.
\newblock In {\em 2009 16th IEEE International Conference on Image Processing
  (ICIP)}, pages 997--1000. IEEE, 2009.

\bibitem{liu2009studying}
Hantao Liu and Ingrid Heynderickx.
\newblock Studying the added value of visual attention in objective image
  quality metrics based on eye movement data.
\newblock In {\em 2009 16th IEEE international conference on image processing
  (ICIP)}, pages 3097--3100. IEEE, 2009.

\bibitem{chen2003visual}
Li-Qun Chen, Xing Xie, Xin Fan, Wei-Ying Ma, Hong-Jiang Zhang, and He-Qin Zhou.
\newblock A visual attention model for adapting images on small displays.
\newblock {\em Multimedia systems}, 9(4):353--364, 2003.

\bibitem{chen2009sketch2photo}
Tao Chen, Ming-Ming Cheng, Ping Tan, Ariel Shamir, and Shi-Min Hu.
\newblock Sketch2photo: Internet image montage.
\newblock {\em ACM transactions on graphics (TOG)}, 28(5):1--10, 2009.

\bibitem{yang2022visual}
Zuyi Yang, Qinghui Dai, and Junsong Zhang.
\newblock Visual perception driven collage synthesis.
\newblock {\em Computational Visual Media}, 8(1):79--91, 2022.

\bibitem{huang2011arcimboldo}
Hua Huang, Lei Zhang, and Hong-Chao Zhang.
\newblock Arcimboldo-like collage using internet images.
\newblock In {\em Proceedings of the 2011 SIGGRAPH Asia Conference}, pages
  1--8, 2011.

\bibitem{smeulders2013visual}
Arnold~WM Smeulders, Dung~M Chu, Rita Cucchiara, Simone Calderara, Afshin
  Dehghan, and Mubarak Shah.
\newblock Visual tracking: An experimental survey.
\newblock {\em IEEE transactions on pattern analysis and machine intelligence},
  36(7):1442--1468, 2013.

\bibitem{pietikainen2011computer}
Matti Pietik{\"a}inen, Abdenour Hadid, Guoying Zhao, and Timo Ahonen.
\newblock {\em Computer vision using local binary patterns}, volume~40.
\newblock Springer Science \& Business Media, 2011.

\bibitem{haidekker2011advanced}
Mark Haidekker.
\newblock {\em Advanced biomedical image analysis}.
\newblock John Wiley \& Sons, 2011.

\bibitem{knutsson1983texture}
H.~Knutsson and G~Granlund.
\newblock Texture analysis using two-dimensional quadrature filters.
\newblock In {\em IEEE Comput. Soc. Workshop on Computer Architecture for
  Pattern Analysis and Image Database Management}, pages 206--213, 1983.

\bibitem{ojala1996comparative}
Timo Ojala, Matti Pietik{\"a}inen, and David Harwood.
\newblock A comparative study of texture measures with classification based on
  featured distributions.
\newblock {\em Pattern recognition}, 29(1):51--59, 1996.

\bibitem{laws1980textured}
Kenneth~I Laws.
\newblock {\em Textured image segmentation}.
\newblock PhD thesis, University of Southern California Los Angeles Image
  Processing INST, 1980.

\bibitem{tan2010enhanced}
Xiaoyang Tan and Bill Triggs.
\newblock Enhanced local texture feature sets for face recognition under
  difficult lighting conditions.
\newblock {\em IEEE transactions on image processing}, 19(6):1635--1650, 2010.

\bibitem{margolin2013makes}
Ran Margolin, Ayellet Tal, and Lihi Zelnik-Manor.
\newblock What makes a patch distinct?
\newblock In {\em Proceedings of the IEEE conference on computer vision and
  pattern recognition}, pages 1139--1146, 2013.

\bibitem{zhang2017salient}
Qing Zhang, Jiajun Lin, Yanyun Tao, Wenju Li, and Yanjiao Shi.
\newblock Salient object detection via color and texture cues.
\newblock {\em Neurocomputing}, 243:35--48, 2017.

\bibitem{porebski2008haralick}
Alice Porebski, Nicolas Vandenbroucke, and Ludovic Macaire.
\newblock Haralick feature extraction from lbp images for color texture
  classification.
\newblock In {\em 2008 First Workshops on Image Processing Theory, Tools and
  Applications}, pages 1--8. IEEE, 2008.

\bibitem{treisman1980feature}
Anne~M Treisman and Garry Gelade.
\newblock A feature-integration theory of attention.
\newblock {\em Cognitive psychology}, 12(1):97--136, 1980.

\bibitem{wolfe2004attributes}
Jeremy~M Wolfe and Todd~S Horowitz.
\newblock What attributes guide the deployment of visual attention and how do
  they do it?
\newblock {\em Nature reviews neuroscience}, 5(6):495--501, 2004.

\bibitem{itti1998model}
Laurent Itti, Christof Koch, and Ernst Niebur.
\newblock A model of saliency-based visual attention for rapid scene analysis.
\newblock {\em IEEE Transactions on pattern analysis and machine intelligence},
  20(11):1254--1259, 1998.

\bibitem{frintrop2015traditional}
Simone Frintrop, Thomas Werner, and German Martin~Garcia.
\newblock Traditional saliency reloaded: A good old model in new shape.
\newblock In {\em Proceedings of the IEEE conference on computer vision and
  pattern recognition}, pages 82--90, 2015.

\bibitem{achanta2008salient}
Radhakrishna Achanta, Francisco Estrada, Patricia Wils, and Sabine
  S{\"u}sstrunk.
\newblock Salient region detection and segmentation.
\newblock In {\em International conference on computer vision systems}, pages
  66--75. Springer, 2008.

\bibitem{cheng2015global}
Ming-Ming Cheng, Niloy~J Mitra, Xiaolei Huang, Philip~HS Torr, and Shi-Min Hu.
\newblock Global contrast based salient region detection.
\newblock {\em IEEE Transactions on Pattern Analysis and Machine Intelligence},
  37(3):569--582, 2015.

\bibitem{guo2009novel}
Chenlei Guo and Liming Zhang.
\newblock A novel multiresolution spatiotemporal saliency detection model and
  its applications in image and video compression.
\newblock {\em IEEE transactions on image processing}, 19(1):185--198, 2009.

\bibitem{perazzi2012saliency}
Federico Perazzi, Philipp Kr{\"a}henb{\"u}hl, Yael Pritch, and Alexander
  Hornung.
\newblock Saliency filters: Contrast based filtering for salient region
  detection.
\newblock In {\em Computer Vision and Pattern Recognition (CVPR), 2012 IEEE
  Conference on}, pages 733--740. IEEE, 2012.

\bibitem{goferman2012context}
Stas Goferman, Lihi Zelnik-Manor, and Ayellet Tal.
\newblock Context-aware saliency detection.
\newblock {\em IEEE transactions on pattern analysis and machine intelligence},
  34(10):1915--1926, 2012.

\bibitem{qi2015saliencyrank}
Wei Qi, Ming-Ming Cheng, Ali Borji, Huchuan Lu, and Lian-Fa Bai.
\newblock Saliencyrank: Two-stage manifold ranking for salient object
  detection.
\newblock {\em Computational Visual Media}, 1(4):309--320, 2015.

\bibitem{borji2015salient}
Ali Borji, Ming-Ming Cheng, Huaizu Jiang, and Jia Li.
\newblock Salient object detection: A benchmark.
\newblock {\em IEEE transactions on image processing}, 24(12):5706--5722, 2015.

\bibitem{maenpaa2004classification}
Topi M{\"a}enp{\"a}{\"a} and Matti Pietik{\"a}inen.
\newblock Classification with color and texture: jointly or separately?
\newblock {\em Pattern recognition}, 37(8):1629--1640, 2004.

\bibitem{achanta2012slic}
Radhakrishna Achanta, Appu Shaji, Kevin Smith, Aurelien Lucchi, Pascal Fua, and
  Sabine S{\"u}sstrunk.
\newblock Slic superpixels compared to state-of-the-art superpixel methods.
\newblock {\em IEEE transactions on pattern analysis and machine intelligence},
  34(11):2274--2282, 2012.

\bibitem{borji2012exploiting}
Ali Borji and Laurent Itti.
\newblock Exploiting local and global patch rarities for saliency detection.
\newblock In {\em 2012 IEEE conference on computer vision and pattern
  recognition}, pages 478--485. IEEE, 2012.

\bibitem{faloutsos1995FastMap}
Christos Faloutsos and King-Ip Lin.
\newblock {\em FastMap: A fast algorithm for indexing, data-mining and
  visualization of traditional and multimedia datasets}, volume~24.
\newblock ACM, 1995.

\bibitem{shi2016hierarchical}
Jianping Shi, Qiong Yan, Li~Xu, and Jiaya Jia.
\newblock Hierarchical image saliency detection on extended cssd.
\newblock {\em IEEE transactions on pattern analysis and machine intelligence},
  38(4):717--729, 2016.

\bibitem{yan2013hierarchical}
Qiong Yan, Li~Xu, Jianping Shi, and Jiaya Jia.
\newblock Hierarchical saliency detection.
\newblock In {\em Proceedings of the IEEE conference on computer vision and
  pattern recognition}, pages 1155--1162, 2013.

\end{thebibliography}

\end{document}